%% file: iclr2027_conference.tex
\documentclass{article}
\usepackage{iclr2027_conference,times}
\input{math_commands.tex}

\usepackage{url}
\usepackage{amsmath,amssymb,amsthm,mathtools}
\usepackage{booktabs}
\usepackage{longtable}
\usepackage{graphicx}
\usepackage{multirow}
\usepackage{array}
\usepackage{enumitem}
\usepackage{xcolor}
\usepackage{algorithm}
\usepackage{algpseudocode}
\usepackage{float}
\usepackage{placeins}
\usepackage{titletoc}
\usepackage{microtype}
\usepackage{hyperref}
\usepackage[nameinlink,capitalise]{cleveref}
\hypersetup{colorlinks=true,linkcolor=blue,citecolor=blue,urlcolor=blue}

\newtheorem{lemma}{Lemma}
\newtheorem{proposition}[lemma]{Proposition}
\newtheorem{theorem}[lemma]{Theorem}

\newcommand{\Prob}{\mathbb{P}}
\newcommand{\sgn}{\operatorname{sign}}
\newcommand{\ind}{\mathbf{1}}
\newcommand{\method}{\textsc{ExPPO}}
\newcommand{\grpoh}{\textsc{GRPO-H}}

\makeatletter
\renewcommand{\paragraph}{\@startsection{paragraph}{4}{\z@}%
  {0.75ex plus 0.2ex minus 0.1ex}%
  {-0.75em}%
  {\normalfont\normalsize\bfseries}}
\makeatother

\title{Exploration-Preserving Policy Optimization}

\author{%
\makebox[\dimexpr\textwidth-2\tabcolsep\relax][c]{%
\begin{tabular}{c}
Hangzhan Jin$^{1,2}$\thanks{Corresponding author: \texttt{hangzhan.jin@mila.quebec}.}
\quad
Mohammad Hamdaqa$^{2}$
\quad
Doina Precup$^{1,3,4,5}$\\[0.35em]
\normalfont\footnotesize
$^{1}$Mila - Quebec AI Institute
\quad
$^{2}$Polytechnique Montr\'eal
\quad
$^{3}$McGill University\\[-0.1em]
\normalfont\footnotesize
$^{4}$CIFAR AI Chair
\quad
$^{5}$Google DeepMind
\end{tabular}%
}
}
\iclrfinalcopy
\hypersetup{
  pdftitle={Exploration-Preserving Policy Optimization},
  pdfauthor={Hangzhan Jin; Mohammad Hamdaqa; Doina Precup},
  pdfsubject={Preprint}
}

\begin{document}
\raggedbottom
\maketitle
\vspace{-18pt} % Reclaim title whitespace for the corresponding-author footnote.
% Override the template's conference-publication header for this preprint.
\fancyhead[L]{Preprint}

\begin{abstract}
\vspace{-6pt} % Keep the added contact and code link within the existing page budget.
Reinforcement learning with verifiable rewards improves reasoning, while the allocation of learning signal shapes which solutions remain accessible under repeated sampling. Group-relative objectives assign equal advantages to equally rewarded responses, making aggregate credit proportional to sampled mode frequency. We introduce Exploration-Preserving Policy Optimization (ExPPO), a lightweight advantage-shaping rule that redistributes credit using prompt-relative, length-normalized response surprisal and prompt pass rate. ExPPO combines bounded shaping with shared normalization to preserve verifier polarity and approximately maintain each prompt group’s total absolute sequence-advantage mass. Our analysis characterizes response-level credit allocation alongside sampled mode updates, deriving local conditions for gains in entropy and correct-mode discovery. Experiments show improved in-domain and out-of-domain reasoning coverage, higher aggregate response accuracy, and strong coverage at large sampling budgets. A controlled multi-answer evaluation further demonstrates increased correct-mode yield and gains in diversity among verified-correct responses. Code: \url{https://github.com/jinhangzhan/ExPPO.git}.
\end{abstract}
\vspace{-6pt}

\section{Introduction}
\label{sec:introduction}

Reinforcement learning with verifiable rewards (RLVR) post-trains language models from automatically checked outcomes and has produced strong gains in mathematical reasoning \citep{guo2025deepseekr1}. At each update, the policy samples several responses for each prompt and learns from their sequence-level verifier rewards. Reasoning systems often use repeated sampling, so their utility depends on both single-sample accuracy and the probability of finding a correct solution across $K$ samples. Recent studies report that RLVR can improve pass@1 while reducing pass@$K$ or answer diversity \citep{yue2025rlcapacity,wu2025invisible}.

Group Relative Policy Optimization (GRPO) is widely used for RLVR because group-normalized rewards replace PPO's learned critic \citep{schulman2017ppo,shao2024deepseekmath}. Equal-reward responses receive equal sequence advantages, so aggregate coefficient mass follows sampled mode frequency. Proposition~\ref{prop:multiplicity-logodds} characterizes the corresponding relative-odds change in a categorical mode model, and finite groups give rare correct modes fewer updates \citep{plyusov2026fgrpo}. GRPO variants share this credit structure \citep{yu2025dapo,zheng2025gspo}, and recent analyses connect it to support concentration and entropy decline \citep{lochab2026ucpo,gai2025smoothing}. \citet{cui2025entropy} connect entropy change to log-probability--advantage covariance and propose Clip-Cov and KL-Cov to restrict high-covariance token updates. These controls target entropy; useful exploration also depends on allocating credit across distinct correct and incorrect responses.

Recent work addresses exploration through entropy control, collection-level objectives, and rarity-aware credit. Direct entropy bonuses have task-dependent benefits in MuJoCo \citep{andrychowicz2021matters,choe2024eapo} and show strong coefficient sensitivity in reasoning LMs \citep{cui2025entropy}. For GRPO-H, the bonus promotes token uncertainty independently of verifier credit (\cref{app:entropy-control}). Entropy methods regulate policy uncertainty; REPO-R uses token-level surprisal to rescale positive and negative advantages toward an entropy target \citep{petrenko2026entropy,zhang2025aer,zhang2026polarity,cheng2026gepo}. Pass@$K$ and set-level methods optimize the collective utility of sampled responses \citep{walder2025pkpo,chen2025passk,orney2026polye}. Rewarding the Unlikely emphasizes low-probability correct responses, F-GRPO scales prompt groups by their success rates, and correct-set methods redistribute credit across successful solutions \citep{he2025unlikely,plyusov2026fgrpo,cao2026cuegrpo}. These approaches motivate prompt-relative control over how each group's learning signal is allocated across responses.

We introduce \textbf{Exploration-Preserving Policy Optimization (\method{})}, a prompt-relative, mass-normalized advantage-shaping rule for group-relative RLVR. \method{} combines group-standardized length-normalized rollout surprisal, a bounded transformation, pass-rate gating, and shared absolute-advantage-mass normalization. The rule preserves verifier polarity and assigns higher signed credit to higher-surprisal responses within each equal-advantage class. Using existing rollout quantities, it integrates into GRPO and DAPO.

We derive sampled mode-logit updates for GRPO and \method{} in a common geometry and characterize response-level allocation with a KL-local model. The analysis establishes polarity preservation and surprisal ordering, together with conditional local results for entropy and correct-mode discovery. Experiments with Qwen2.5-7B, Qwen3-0.6B, and Phi-3.5-mini measure response accuracy, large-$K$ coverage, entropy dynamics, and diversity among verifier-checked answer modes.

Our contributions are threefold:
\begin{itemize}[leftmargin=1.35em,itemsep=2pt,topsep=3pt]
    \item We characterize sampled mode-level credit allocation in group-relative RLVR and its dependence on response multiplicity.
    \item Our method, \method{}, combines bounded response surprisal, pass-rate gating, and absolute-advantage-mass normalization in a prompt-relative redistribution rule.
    \item We evaluate accuracy, coverage, and correct-mode diversity, including closest-baseline comparisons and independently replicated Phi-3.5-mini training.
\end{itemize}

\section{Preliminaries}
\label{sec:preliminaries}

\paragraph{Reinforcement learning with verifiable rewards.}
For a prompt $q$ from training distribution $\mathcal D$, a frozen rollout policy $\pi_{\mathrm{old}}$ independently samples $G$ responses. We use $Y_i$ for a random response and $y_i=(y_{i,1},\ldots,y_{i,T_i})$ for its realized value, with length $T_i$ and token context $x_{i,t}=(q,y_{i,<t})$. A verifier assigns reward $r_i=R(q,y_i)$. The current policy $\pi_\theta$ learns from these sequence rewards.

\paragraph{Group-relative policy optimization.}
GRPO assigns the detached sequence advantage
\begin{equation}
A_i=\frac{r_i-\mu_{r,q}}{\sigma_{r,q}+\varepsilon},\qquad
\mu_{r,q}=\frac1G\sum_{j=1}^{G}r_j,\qquad
\sigma_{r,q}^2=\frac1G\sum_{j=1}^{G}(r_j-\mu_{r,q})^2,
\label{eq:grpo-advantage}
\end{equation}
where $\varepsilon>0$ is a numerical stabilizer \citep{shao2024deepseekmath}. The same $A_i$ applies to every token in $y_i$. Writing $\mathcal B$ for prompt/group sampling and $\langle\cdot\rangle$ for token averaging, GRPO and its entropy-regularized baseline, \grpoh{}, maximize
\begin{align}
\mathcal{J}_{\mathrm{GRPO}}(\theta)
&=\mathbb{E}_{\mathcal{B}}\!\left[\left\langle\ell_{\mathrm{clip}}(w_{i,t}(\theta),A_i)-\beta_{\mathrm{KL}}d_{i,t}(\theta)\right\rangle\right],
\label{eq:grpo-objective}\\
\mathcal{J}_{\grpoh{}}(\theta)
&=\mathcal{J}_{\mathrm{GRPO}}(\theta)+\beta_{\mathrm{ent}}\mathbb{E}_{\mathcal{B}}\!\left[\left\langle H(\pi_\theta(\cdot\mid x_{i,t}))\right\rangle\right].
\label{eq:entropy-bonus}
\end{align}
Here $w_{i,t}$ is the token importance ratio, $\ell_{\mathrm{clip}}$ the clipped surrogate, $d_{i,t}$ the reference-policy KL, and $H$ Shannon entropy \citep{schulman2017ppo,mnih2016a3c}. Details appear in \cref{app:grpo-details}.

\section{From Reward Credit to Support Allocation}
\label{sec:motivation}

This section examines sampled mode-level coefficients and motivates the allocation rule in \cref{sec:method}.

\paragraph{Equal reward credit gives multiplicity-weighted coefficients.}
Under binary rewards, every correct response in a mixed group receives the same positive GRPO advantage $A^{+}$, and every incorrect response receives the same negative value $A^{-}$; the closed form is given in \cref{app:binary-grpo}. Suppose a correct response mode $m$ appears $n_m$ times. Its aggregate positive coefficient is
\begin{equation}
    C_m=\sum_{i:y_i\in m}A_i=n_mA^{+}.
    \label{eq:mode-credit}
\end{equation}
Equal per-response credit therefore produces aggregate mode-level credit proportional to sample multiplicity. The categorical mode update in Proposition~\ref{prop:multiplicity-logodds} converts a count difference into the relative log-odds shift $\eta A^{+}(n_m-n_{m'})$ between two correct modes $m$ and $m'$. Recurring failure modes receive the same exposure effect on the negative side. Finite-group analyses show that rare-correct trajectories also enter fewer updates \citep{plyusov2026fgrpo}. 

\paragraph{Global entropy combines correct and failure diversity.}\label{sec:entropy-motivation}
PPO/TRPO-style entropy bonuses reward uncertainty at sampled states; MuJoCo studies report task-dependent gains and benefits from trajectory-level entropy credit \citep{andrychowicz2021matters,choe2024eapo}. For \grpoh{}, the bonus favors unlikely tokens regardless of correctness, while equal-reward responses retain equal GRPO advantages. Entropy can therefore grow through failed continuations (\cref{app:entropy-control}). To connect this signal to correct support, let $Y\sim\pi(\cdot\mid q)$ be a random response and $C=R(q,Y)\in\{0,1\}$ its verifier-defined success indicator. Writing $p_q=\Prob(C=1\mid q)$, the entropy chain rule gives
\begin{equation}
H(Y\mid q)=H_b(p_q)+p_qH(Y\mid C=1,q)+(1-p_q)H(Y\mid C=0,q).
\label{eq:entropy-decomposition}
\end{equation}
where $H_b$ is binary entropy. The three terms measure uncertainty over success, diversity among correct responses, and diversity among failures.

The decomposition shows that global entropy combines these three sources of policy spread. Consider two four-mode policies with success probability $0.5$. One assigns correct masses $\{0.49,0.01\}$ and failure masses $\{0.25,0.25\}$; the other swaps these pairs. The policies share global entropy and total correct mass; the second distributes correct probability evenly across both valid modes. Correct-mode diversity therefore complements global entropy.

\section{Exploration-Preserving Policy Optimization}
\label{sec:method}

\method{} adds a detached correction to the verifier advantage: compare response surprisal within each prompt, use pass rate to allocate the correction, and normalize the shaped advantages. \Cref{tab:quadrants} summarizes the resulting credit allocation.

\paragraph{Prompt-relative response surprisal.}
For response $y_i=(y_{i,1},\ldots,y_{i,T_i})$, we use its length-normalized surprisal under the rollout policy,
\begin{equation}
s_i=-T_i^{-1}\log\pi_{\mathrm{old}}(y_i\mid q),\qquad u_i=\tanh\!\left(\frac{s_i-\mu_{s,q}}{\max(\sigma_{s,q},\varepsilon)}\right).
\label{eq:rarity}
\end{equation}
where $\mu_{s,q}$ and $\sigma_{s,q}$ are computed across the $G$ responses for prompt $q$. The score $u_i\in(-1,1)$ indicates whether length-normalized surprisal is above or below the group mean. We use \emph{response rarity} as shorthand for this policy-relative statistic; the controlled evaluation defines semantic modes through exact answer identity. Length normalization controls length effects; group standardization puts responses on a prompt-relative scale; $\tanh$ bounds extreme scores. Rollout-policy quantities are detached.

\paragraph{Pass-rate-gated advantage shaping.}
Sampled prompt pass rate determines the strength assigned to the surprisal signal. Let $c_i=\ind[r_i>0]$ denote verifier-defined success and $\rho_q=G^{-1}\sum_{j=1}^{G}c_j$ the group pass rate. Graded rewards retain the original standardized advantage $A_i$ and use a predeclared success threshold. The gate $g_i$ and coefficient $\lambda\in[0,1]$ define
\begin{equation}
g_i=(1-\rho_q)\ind[u_i>0]+\rho_q\ind[u_i<0].
\label{eq:pass-rate-gate}
\end{equation}
\begin{equation}
\widehat A_i=A_i+\lambda|A_i|g_i u_i,\qquad 0\leq\lambda\leq1.
\label{eq:exppo-shaping}
\end{equation}
The factor $|A_i|$ ties correction strength to verifier credit. Together, $|u_i|<1$ and $\lambda\leq1$ preserve each advantage's sign. Low pass rates emphasize rare alternatives; high pass rates reduce repeated reinforcement and strengthen penalties on common errors (\cref{tab:quadrants}).

\begin{table}[!htb]
\centering
\caption{Response-level credit allocation before shared normalization. Rarity is prompt-relative. The gate emphasizes rare responses in low-success groups and common responses in high-success groups; bounded shaping preserves the verifier sign.}
\label{tab:quadrants}
\small
\setlength{\tabcolsep}{4pt}
\begin{tabular*}{\linewidth}{@{\extracolsep{\fill}}llll@{}}
\toprule
Response & $A_i>0$: reinforcement & $A_i<0$: penalty & Gate $g_i$ \\
\midrule
Rare ($u_i>0$) & Stronger & Softer & $1-\rho_q$ \\
Common ($u_i<0$) & Weaker & Stronger & $\rho_q$ \\
\bottomrule
\end{tabular*}
\end{table}

Shared normalization keeps the prompt group's total credit comparable to the base rule. We use the absolute mass so reinforcement and penalties both count toward this budget:
\begin{equation}
\alpha_q=\frac{\sum_{j=1}^{G}|A_j|}{\sum_{j=1}^{G}|\widehat A_j|+\varepsilon},\qquad \widetilde A_i=\alpha_q\widehat A_i.
\label{eq:normalization}
\end{equation}
with $\widetilde A_i=0$ when $\sum_{j=1}^{G}|A_j|=0$. The shared positive factor preserves response ordering and approximately matches the base absolute sequence-advantage mass. Shaping reallocates credit both within and between reward classes; token aggregation determines the corresponding loss weights.

\paragraph{Integration with group-relative RLVR.}\label{sec:integration}
\method{} substitutes $\widetilde A_i$ for $A_i$ in \cref{eq:grpo-objective}. Its DAPO instantiation retains asymmetric clipping, dynamic sampling, token-level aggregation, and overlong-response treatment \citep{yu2025dapo}. The method reuses rollout log-probabilities, preserves the model set and rollout count, and adds $O(G)$ prompt-group reductions. See \cref{app:implementation,app:overhead}.

\section{Analysis}
\label{sec:analysis}

We first establish polarity and ordering of the shaped coefficients, then connect their allocation to entropy and correct-mode discovery in a KL-local model. Throughout, $Y$ denotes a random response and $y$ a realized value; indexed coefficients such as $A_i$ and $B_i$ are evaluated on $y_i$. A notation key appears in \cref{app:notation}, with proofs in \cref{app:proofs}.

\paragraph{Local response-space update.}
Fix a prompt and its rollout-group statistics, and extend \cref{eq:rarity,eq:exppo-shaping,eq:normalization} to responses in the rollout support using these frozen statistics. The score $h=\widetilde A$ describes response-level advantage allocation. A parameter update direction $d$ induces the score $h_d(y)=\nabla_\theta\log\pi_\theta(y\mid q)^\top d$, evaluated at $\theta=\theta_{\mathrm{old}}$, whose exponential tilt matches the parameter update's first-order density change. See \cref{app:binary-grpo}.

\begin{lemma}[KL-regularized local update]
\label{lem:tilt}
For step size $\eta>0$, the solution of
\begin{equation}
\max_{\pi}\left\{\E_{Y\sim\pi}[h(q,Y)]-\eta^{-1}D_{\mathrm{KL}}(\pi\|\pi_{\mathrm{old}})\right\}.
\label{eq:local-kl}
\end{equation}
is the exponential tilt
\begin{equation}
\pi_{\eta,h}(y\mid q)=\frac{\pi_{\mathrm{old}}(y\mid q)e^{\eta h(q,y)}}{\E_{Y'\sim\pi_{\mathrm{old}}}[e^{\eta h(q,Y')}]}. 
\label{eq:tilt}
\end{equation}
and $\left.\frac{d}{d\eta}\E_{\pi_{\eta,h}}[V]\right|_{\eta=0}=\Cov_{\pi_{\mathrm{old}}}(V,h)$ for any fixed response utility $V$.
\end{lemma}

This exponential-tilt model follows the KL geometry of natural-gradient and trust-region policy optimization and makes relative probability changes explicit \citep{kakade2001natural,schulman2015trpo}.

\paragraph{Bounded polarity-preserving redistribution.}
Write $B_i=|A_i|g_i u_i$, so that $\widehat A_i=A_i+\lambda B_i$. Equivalently, $\widehat A_i=A_iM_i$ with $M_i=1+\lambda g_i\sgn(A_i)u_i$.

\begin{proposition}[Bounded reward-polarity preservation]
\label[proposition]{prop:bounded}
For $0\leq\lambda\leq1$, $g_i\in[0,1]$, and $u_i\in[-1,1]$,
\begin{equation}
1-\lambda\leq M_i\leq1+\lambda,\qquad A_i\widehat A_i\geq0,\qquad |\widehat A_i-A_i|\leq\lambda|A_i|.
\label{eq:bounded-main}
\end{equation}
The normalized advantage $\widetilde A_i$ has the same sign as $\widehat A_i$. For $\lambda<1$, every nonzero base advantage strictly preserves its sign.
\end{proposition}

Every nonzero shaped advantage retains the sign of its base verifier advantage. Relative surprisal changes the response's share of the group coefficients. The intervention ratio $\kappa(\lambda)=(1+\lambda)/(1-\lambda)$ bounds relative multiplier spread for $\lambda<1$ (\cref{app:proofs}).

\paragraph{Surprisal ordering and response-space odds.}
The ordering result concerns responses that receive equal credit under the base objective. For a generic sampled pass rate $\rho\in(0,1)$, define
\begin{equation}
\psi_{\rho}(u)=u\big((1-\rho)\ind[u>0]+\rho\ind[u<0]\big).
\label{eq:psi}
\end{equation}

We write $\pi_{\eta,\widetilde A}$ for the exponential tilt in \cref{eq:tilt} with response score $h(q,y_i)=\widetilde A_i$ on the group.

\begin{theorem}[Surprisal-ordered response preference]
\label[theorem]{thm:rare-preference}
Fix a mixed group with $0<\rho_q<1$, a nonzero base advantage $A$, and $\lambda>0$. The score $h_\lambda(u)=A+\lambda|A|\psi_{\rho_q}(u)$ is strictly increasing in $u$. For any response policy $\pi$, define $\Delta_{ij}(\pi)=\log\pi(y_i\mid q)-\log\pi(y_j\mid q)$. Hence, two sampled responses with $A_i=A_j$ and $u_i>u_j$ satisfy $\widetilde A_i>\widetilde A_j$ and
\begin{equation}
\Delta_{ij}(\pi_{\eta,\widetilde A})-\Delta_{ij}(\pi_{\mathrm{old}})=\eta(\widetilde A_i-\widetilde A_j)>0.
\label{eq:relative-odds}
\end{equation}
\end{theorem}

Within this KL-local family, the base score $A$ preserves equal-credit odds, and \method{} shifts those odds toward the higher-surprisal response. The sampled logit counterpart appears in \cref{app:binary-grpo}.

\paragraph{Surprisal alignment, entropy, and matched-KL utility.}
The pairwise theorem describes one equal-credit class. At the group level, the correction is aligned with the implemented surprisal statistic. We write $\widehat{\Cov}_q$ for covariance under the uniform distribution over the realized group.

\begin{proposition}[Empirical surprisal alignment]
\label[proposition]{prop:rarity-alignment}
Under the uniform distribution over a prompt group,
\begin{equation}
\widehat{\Cov}_q(s,B)=G^{-1}\sum_{i=1}^{G}(s_i-\mu_{s,q})B_i\geq0.
\label{eq:rarity-covariance}
\end{equation}
\end{proposition}

To connect this direction to Shannon entropy, freeze the group statistics and draw $Y\sim\pi_{\mathrm{old}}(\cdot\mid q)$. Write $A=A(Y)$, $B=B(Y)$, and $S=-\log\pi_{\mathrm{old}}(Y\mid q)$ for the resulting random variables. Unmarked covariances and variances use this response distribution. For $h=\alpha_q(A+\lambda B)$ with fixed $\alpha_q>0$, \cref{lem:tilt} gives
\begin{equation}
\left.\frac{dH(\pi_{\eta,h}(\cdot\mid q))}{d\eta}\right|_{0}=\alpha_q\big[\Cov(S,A)+\lambda\Cov(S,B)\big].
\label{eq:entropy-slope}
\end{equation}
The first term is the entropy direction induced by the reward advantage; the second gives \method{}'s additional contribution. Proposition~\ref{prop:rarity-alignment} establishes empirical alignment with length-normalized surprisal $s$, and \cref{eq:entropy-slope} characterizes response entropy through full-response surprisal $S$. Logged next-token entropy in \cref{sec:entropy-results} provides a complementary training diagnostic.

Correct-mode discovery gives a concrete exploration utility. For correct modes $m\in\mathcal M_q^+$ with masses $p_m(\pi)=\Prob_{Y\sim\pi}(Y\in m\mid q)$, the expected number observed in $K$ independent responses is $F_K(\pi)=\sum_m[1-(1-p_m(\pi))^K]$. Its local response utility is $V_K(y)=K(1-p_{m(y)}(\pi_{\mathrm{old}}))^{K-1}$ for correct $y$ and zero for failures. For $K\geq2$, this utility gives greater weight to low-mass correct modes. See \cref{app:proofs}. For a fixed utility $V$, write $J_V(\pi)=\mathbb{E}_{Y\sim\pi}[V(Y)]$. For a small KL budget $\delta>0$, let $\eta_h=\eta_h(\delta)$ satisfy $D_{\mathrm{KL}}(\pi_{\eta_h,h}\|\pi_{\mathrm{old}})=\delta$ and define
\begin{equation}
    \Gamma_V(h)=\frac{\Cov(V,h)}{\sqrt{\Var(h)}}.
    \label{eq:gamma}
\end{equation}
At a common KL budget, the local gain is $J_V(\pi_{\eta_h,h})-J_V(\pi_{\mathrm{old}})=\sqrt{2\delta}\,\Gamma_V(h)+O(\delta)$, where $O(\delta)$ collects higher-order terms. Assuming $\Var(A)>0$, define the component of $B$ orthogonal to $A$ as
\begin{equation}
B_\perp=B-\frac{\Cov(A,B)}{\Var(A)}A,\qquad \Cov(A,B_\perp)=0.
\label{eq:b-perp}
\end{equation}
Its alignment with $V$ determines the gain:
\begin{equation}
\left.\frac{d}{d\lambda}\Gamma_V(A+\lambda B)\right|_{\lambda=0}=\frac{\Cov(V,B_\perp)}{\sqrt{\Var(A)}}.
\label{eq:matched-kl-derivative}
\end{equation}
For $V=V_K$, positive residual alignment increases the leading matched-KL gain in expected correct-mode yield for sufficiently small positive $\lambda$. Under binary rewards, $A$ is affine in correctness: $V_1(Y)=C$ gives $\Cov(V_1,B_\perp)=0$, while $K\geq2$ introduces variation across correct modes. Verifier coverage depends on total correct mass through $1-(1-p_q)^K$; mode yield and correct-pair diversity characterize its allocation across correct answers. See \cref{app:finite-lambda}.

\section{Experiments}
\label{sec:experiments}

We evaluate whether response-level rarity shaping preserves repeated-sample coverage, sustains response accuracy, and discovers distinct correct modes.

\subsection{Experimental Setup}
\label{sec:exp-setup}

\paragraph{Eight benchmarks including four In-Domain (ID) and four Out-of-Domain (OOD) for evaluation.} 
Training uses the DAPO-Math-17k RL prompt set and its rule-based verifier \citep{yu2025dapo}. The ID suite comprises AIME24, AIME25, Minerva \citep{lewkowycz2022minerva}, and OlympiadBench \citep{he2024olympiadbench}. The OOD suite comprises GPQA \citep{rein2023gpqa}, MMLU-Pro \citep{wang2024mmlupro}, a safety set, and a four-choice TruthfulQA-MC construction \citep{lin2022truthfulqa}, adapting the ID/OOD evaluation setup of \citet{jin2025oodforgetting,jin2026rpsft}. We call pass@$k$, the probability of at least one verified success in $k$ responses, \emph{coverage}. High-$k$ curves estimate it from the first $k$ generated responses per prompt. We use $k=16$ for AIME24/25 (30 prompts each), $k=8$ for Minerva (272 prompts), and $k=4$ for the other five benchmarks. Response accuracy is the mean verified-response share. Both metrics use unweighted suite averages; see \cref{app:experimental-details}.
\paragraph{Three backbones span scale and checkpoint type.}
We study \textbf{Qwen2.5-7B}, \textbf{Qwen3-0.6B}, and \textbf{Phi-3.5-mini} \citep{qwen2024qwen25,qwen2025qwen3,abdin2024phi3}. Core methods are \textbf{GRPO} \citep{shao2024deepseekmath}, entropy-regularized \textbf{GRPO-H} (\cref{eq:entropy-bonus}), and \textbf{ExPPO}. Focused comparisons add Rewarding the Unlikely (\textbf{GRPO-UR}), \textbf{REPO-R}, and \textbf{F-GRPO} \citep{he2025unlikely,petrenko2026entropy,plyusov2026fgrpo}. See \cref{app:experimental-details}.

\paragraph{Matched budgets support controlled comparisons.} 
The main comparisons share data, optimizer, clipping, token budget, and a 900-step schedule with $G=16$ rollouts per prompt; evaluation uses step 900. The Phi-3.5-mini comparison of F-GRPO, GRPO-UR, REPO-R, and ExPPO uses 300 matched steps and three independent training runs per method. Standard evaluation uses temperature 0.7 and top-$p=0.95$.
\subsection{Performance During Training}
\label{sec:main-results}

\begin{figure}[!htbp]
    \centering
    \setlength{\abovecaptionskip}{4pt}
    \includegraphics[width=0.97\textwidth]{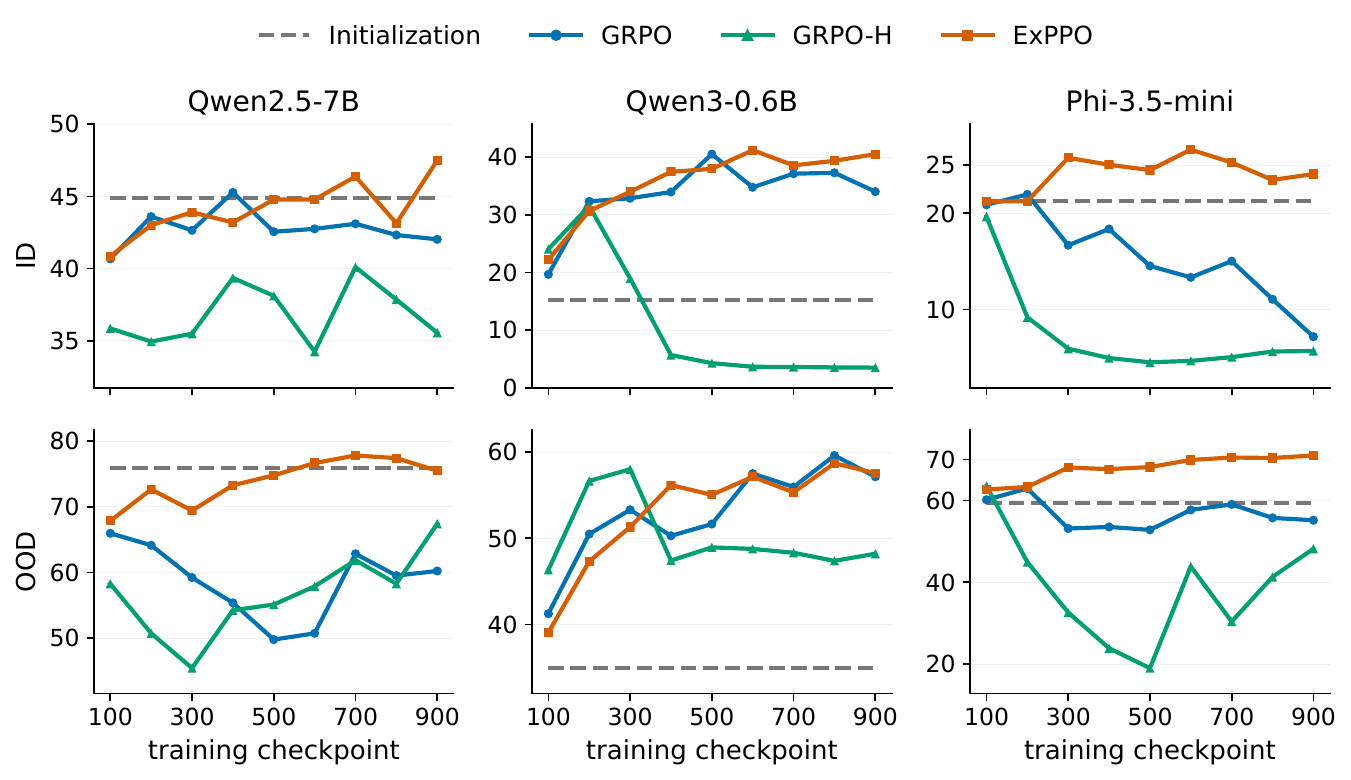}
    \caption{Checkpoint-wise any-success coverage. Columns: Qwen2.5-7B, Qwen3-0.6B, Phi-3.5-mini; rows: ID and OOD four-task means (percentage points). Budgets: $k=16$ for AIME24/25, 8 for Minerva, 4 otherwise. Dashed lines: initialization. Per-task trajectories: \cref{app:per-benchmark-trends}.}
    \label{fig:training-macro}
\end{figure}

\paragraph{ExPPO leads both ID and OOD across all backbones.}
ExPPO finishes highest across all six model--suite trajectories (\cref{fig:training-macro}). It is the only core method to exceed base-model OOD coverage on every backbone during training; GRPO and GRPO-H show OOD forgetting on Qwen2.5-7B and Phi-3.5-mini. This is consistent with exploration supporting knowledge retention during adaptation. OOD representation probes appear in \cref{app:ood-representations}.

\subsection{Entropy Dynamics and Large-Sample Coverage}

\label{sec:entropy-results}

\begin{figure}[!htbp]
    \centering
    \setlength{\abovecaptionskip}{4pt}
    \includegraphics[width=0.97\textwidth]{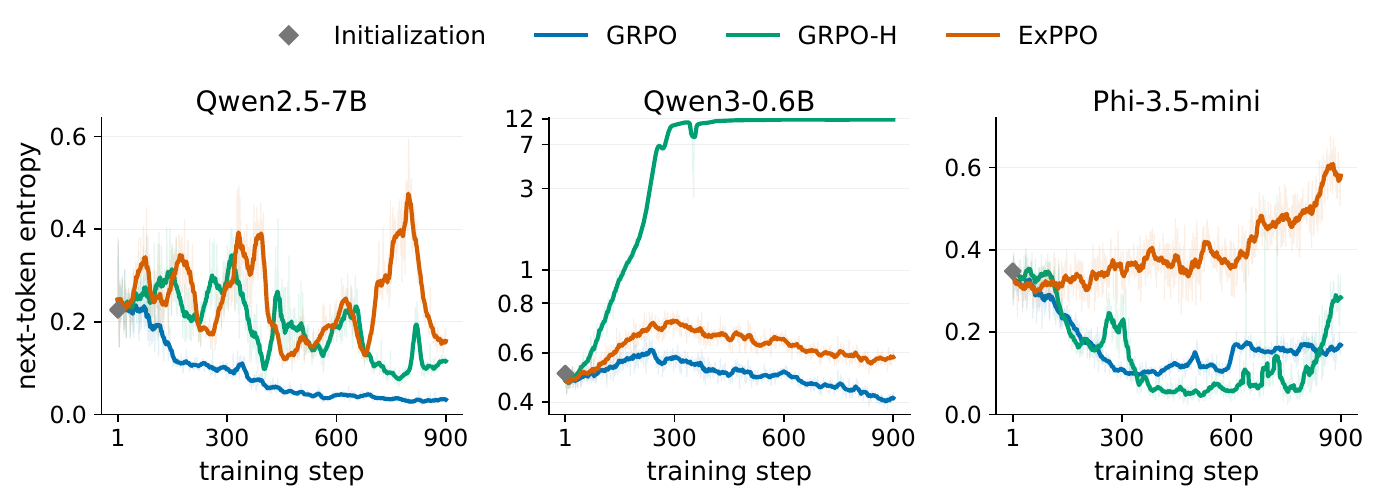}
    \caption{Next-token entropy through training. Faint/solid traces: per-step observations/step means; diamonds: initial states. Qwen3-0.6B uses a linear scale through 0.8 and continuous compression above 0.8, with ticks in entropy units. The other panels use model-specific linear ranges.}
    \label{fig:entropy-dynamics}
\end{figure}

\begin{figure}[!htbp]
    \centering
    % The supplied high-K figure contains initialization, GRPO, GRPO-H, and ExPPO.
    % Insert completed closest-baseline high-K outputs here when available.
    \includegraphics[width=\textwidth]{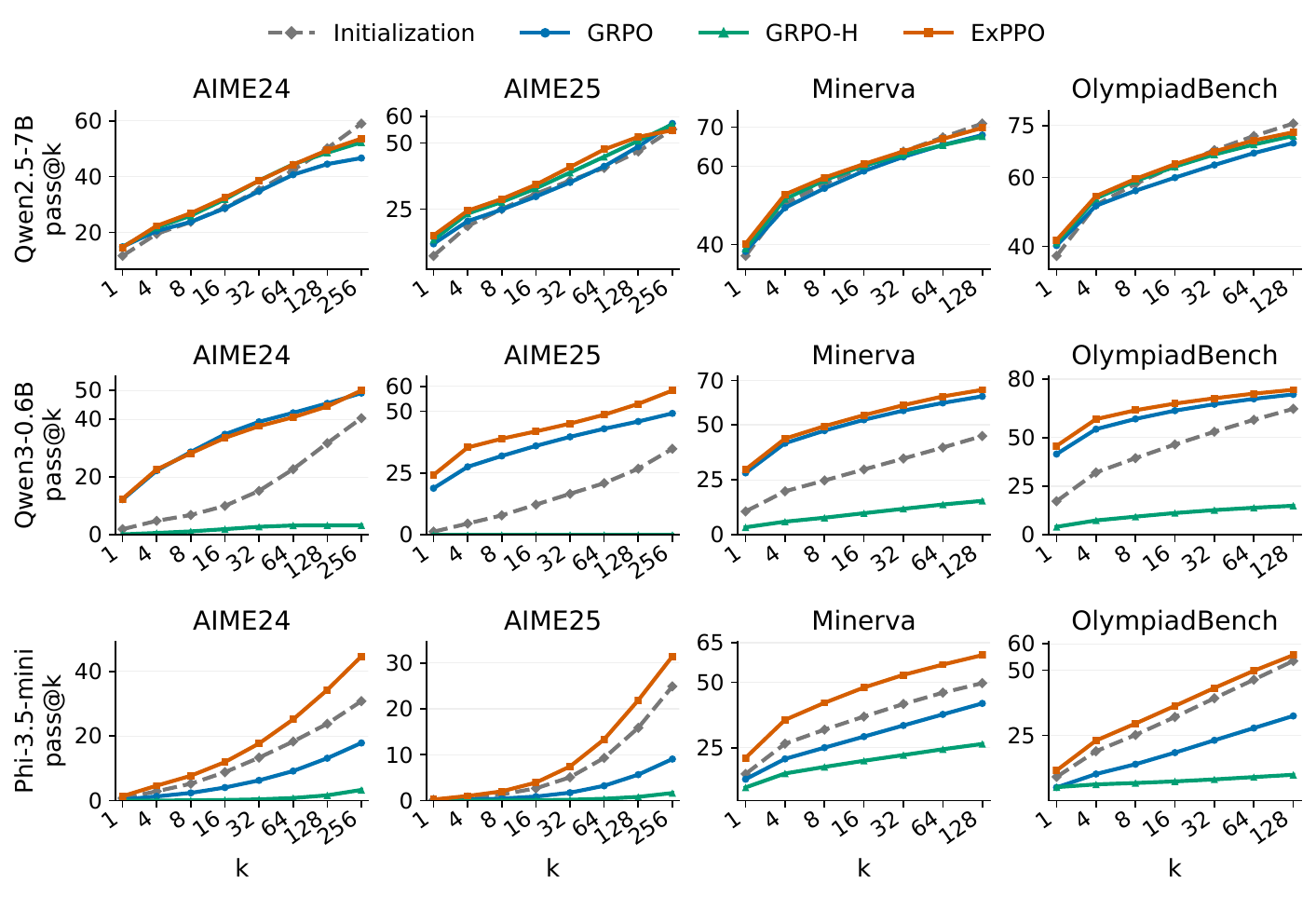}
    \caption{Large-$K$ coverage at step 900, using the first $k$ responses per prompt. Rows: the three backbones; columns: four ID tasks. Curves include initialization and the core training rules. At the largest budget, ExPPO matches or exceeds the other trained core methods on both smaller backbones and three Qwen2.5-7B tasks.}
    \label{fig:high-k}
\end{figure}

\paragraph{Entropy dynamics accompany coverage gains.}
ExPPO retains higher entropy than GRPO on three models (\cref{fig:entropy-dynamics}) with stronger coverage, while GRPO-H combines high entropy with low ID coverage. Qwen2.5-7B initialization remains competitive at the largest budgets (\cref{fig:high-k}).

% Let the controlled-task discussion fill space beside the preceding floats.
\subsection{Controlled Multi-Answer Evaluation}
\label{sec:controlled}

We use 64 synthetic coin-count prompts, each asking for four bounded integer counts whose total coin value matches a target. Each distinct verified count vector defines a correct mode, so measured diversity reflects different valid answers.

\begin{table}[!htb]
\caption{Last-checkpoint controlled evaluation ($K=128$; 64 prompts). A: Accuracy/Solved: correct-response share (0--100)/successful prompts. Observed/effective yields: mean distinct correct modes/exponentiated mode entropy (unsolved: zero). Correct/prompt: successes/64. B: Shared: prompts with $\geq2$ successes per method; $D$: correct-pair mode disagreement averaged on Shared (0--100); $\Delta D$: ExPPO minus baseline; CI: 95\% paired prompt-bootstrap interval. Qwen2.5-7B: updated ExPPO for A and F-GRPO/REPO-R; earlier for GRPO/GRPO-H (\cref{app:multi-answer-details}).}
\label{tab:multi-answer}
\centering
\footnotesize
\renewcommand{\arraystretch}{0.95}
\setlength{\tabcolsep}{3.8pt}
\begin{tabular*}{\linewidth}{@{\extracolsep{\fill}}llrrrrr@{}}
\toprule
\multicolumn{7}{@{}l}{\textbf{A. Fixed-budget correctness and correct-mode yield}} \\
Model & Method & Accuracy & Solved / 64 & \shortstack[r]{Observed yield} & \shortstack[r]{Effective yield} & \shortstack[r]{Correct / prompt} \\
\midrule
\multirow{5}{*}{Qwen2.5-7B}
& GRPO & 6.45 & 57 & 1.750 & 1.546 & 8.25 \\
& GRPO-H & 9.68 & 59 & 1.906 & 1.610 & 12.39 \\
& F-GRPO & \textbf{35.51} & \textbf{64} & 3.234 & 2.389 & \textbf{45.45} \\
& REPO-R & 15.42 & \textbf{64} & 3.578 & 2.820 & 19.73 \\
& \textbf{ExPPO} & 33.50 & \textbf{64} & \textbf{4.172} & \textbf{3.013} & 42.88 \\
\addlinespace[2pt]
\multirow{3}{*}{Qwen3-0.6B}
& GRPO   & \textbf{98.43} & \textbf{64} & 5.422 & 3.013 & \textbf{125.98} \\
& GRPO-H & 0.02  & 1  & 0.016 & 0.016 & 0.03 \\
& \textbf{ExPPO} & 97.74 & \textbf{64} & \textbf{6.016} & \textbf{3.481} & 125.11 \\
\addlinespace[2pt]
\multirow{3}{*}{Phi-3.5-mini}
& GRPO   & 0.90 & 22 & 0.438 & 0.407 & 1.16 \\
& GRPO-H & 0.50 & 9  & 0.219 & 0.206 & 0.64 \\
& \textbf{ExPPO} & \textbf{2.26} & \textbf{33} & \textbf{0.703} & \textbf{0.634} & \textbf{2.89} \\
\bottomrule
\end{tabular*}
\vspace{3pt}
%\begin{tabular*}{\linewidth}{@{\extracolsep{\fill}}llrrrrr@{}}
%\toprule
%\multicolumn{7}{@{}l}{\textbf{B. Conditional correct-pair diversity}} \\
%Model & Baseline & Shared / 64 & Baseline $D$ & ExPPO $D$ & $\Delta D$ & CI \\
%\midrule
%Qwen2.5-7B & GRPO & 51 & 33.98 & \textbf{50.57} & \textbf{$+16.59$} & \textbf{$[+6.19,+27.10]$} \\
%            & GRPO-H & 55 & 35.14 & \textbf{48.77} & \textbf{$+13.63$} & \textbf{$[+5.37,+21.71]$} \\
%            & F-GRPO & 63 & 46.53 & \textbf{56.28} & \textbf{$+9.74$} & \textbf{$[+3.03,+16.51]$} \\
%            & REPO-R & 63 & 56.12 & \textbf{56.28} & $+0.16$ & $[-7.58,+7.78]$ \\
%Qwen3-0.6B & GRPO   & 64 & 51.46 & \textbf{58.44} & \textbf{$+6.98$} & \textbf{$[+3.01,+11.20]$} \\
%\bottomrule
%\end{tabular*}
\begin{tabular*}{\linewidth}{@{\extracolsep{\fill}}llrrrrr@{}}
\toprule
\multicolumn{7}{@{}l}{\textbf{B. Conditional correct-pair diversity}} \\
Model & Baseline & Shared / 64 & Baseline $D$ & ExPPO $D$ & $\Delta D$ & \\
\midrule
Qwen2.5-7B & GRPO & 51 & 33.98 & \textbf{50.57} & \textbf{$+16.59$} & \\
            & GRPO-H & 55 & 35.14 & \textbf{48.77} & \textbf{$+13.63$} &  \\
            & F-GRPO & 63 & 46.53 & \textbf{56.28} & \textbf{$+9.74$} &  \\
            & REPO-R & 63 & 56.12 & \textbf{56.28} & $+0.16$ &  \\
Qwen3-0.6B & GRPO   & 64 & 51.46 & \textbf{58.44} & \textbf{$+6.98$} &  \\
\bottomrule
\end{tabular*}
\end{table}

\paragraph{ExPPO increases fixed-budget mode yield and conditional diversity.}
ExPPO increases observed and effective correct-mode yields over GRPO and GRPO-H across backbones (\cref{tab:multi-answer}). On Qwen2.5-7B, it also broadens conditional diversity relative to both core baselines and F-GRPO, despite F-GRPO's higher accuracy. REPO-R has a similar diversity point estimate and lower yields. Qwen3-0.6B gains diversity at near-saturated accuracy. Full comparisons appear in \cref{app:multi-answer-details}.

\subsection{Comparison with the Closest Baselines}
\label{sec:nearest}
\begin{table}[!htb]
\caption{Qwen2.5-7B last-checkpoint coverage (percentage points): mean $\pm$ sample standard deviation across three decoding seeds at one fixed step-900 checkpoint per method. Initialization: one evaluation. Budgets: \cref{sec:exp-setup}; Avg.: within-run four-task mean. Bold: best trained mean.}
\label{tab:nearest}
\centering
\small
\renewcommand{\arraystretch}{0.95}
\setlength{\tabcolsep}{3.0pt}
\begin{tabular}{@{}clrrrrr@{}}
\toprule
Domain & Method & AIME24 & AIME25 & Minerva & Olympiad & Avg. \\
\midrule
\multirow{7}{*}{ID}
& Initialization & 36.70 & 36.70 & 55.10 & 51.00 & 44.88 \\
\cmidrule(l){2-7}
& GRPO & $23.33{\pm}3.33$ & $24.44{\pm}9.62$ & $53.55{\pm}0.77$ & $50.96{\pm}0.90$ & $38.07{\pm}2.74$ \\
& GRPO-H & $26.67{\pm}10.00$ & $27.78{\pm}10.18$ & $56.50{\pm}1.39$ & $53.28{\pm}1.13$ & $41.06{\pm}4.76$ \\
& GRPO-UR & $25.56{\pm}1.92$ & $21.11{\pm}1.92$ & $56.74{\pm}1.49$ & $47.95{\pm}0.87$ & $37.84{\pm}0.88$ \\
& REPO-R & $26.67{\pm}3.33$ & $33.33{\pm}3.33$ & $57.60{\pm}1.29$ & $\mathbf{56.20{\pm}1.05}$ & $43.45{\pm}0.90$ \\
& F-GRPO & $27.78{\pm}3.85$ & $\mathbf{36.67{\pm}3.33}$ & $56.37{\pm}0.93$ & $53.83{\pm}0.34$ & $43.66{\pm}0.67$ \\
& ExPPO & $\mathbf{34.44{\pm}5.09}$ & $\mathbf{36.67{\pm}3.33}$ & $\mathbf{58.46{\pm}1.33}$ & $55.01{\pm}0.34$ & $\mathbf{46.14{\pm}1.94}$ \\
\midrule
Domain & Method & GPQA & MMLU-Pro & Safety & TruthfulQA & Avg. \\
\midrule
\multirow{7}{*}{OOD}
& Initialization & 59.80 & 81.40 & 84.10 & 78.50 & 75.95 \\
\cmidrule(l){2-7}
& GRPO & $57.66{\pm}3.29$ & $77.62{\pm}0.82$ & $32.30{\pm}0.38$ & $70.57{\pm}0.61$ & $59.54{\pm}0.88$ \\
& GRPO-H & $56.99{\pm}1.23$ & $81.43{\pm}1.43$ & $52.03{\pm}0.47$ & $\mathbf{80.17{\pm}1.08}$ & $67.66{\pm}0.57$ \\
& GRPO-UR & $58.11{\pm}2.17$ & $80.95{\pm}2.18$ & $81.27{\pm}0.55$ & $74.27{\pm}0.67$ & $73.65{\pm}1.30$ \\
& REPO-R & $55.73{\pm}1.58$ & $80.95{\pm}0.82$ & $34.17{\pm}0.49$ & $77.83{\pm}0.55$ & $62.17{\pm}0.18$ \\
& F-GRPO & $57.29{\pm}0.52$ & $77.62{\pm}0.82$ & $83.00{\pm}0.20$ & $79.78{\pm}0.72$ & $74.42{\pm}0.42$ \\
& ExPPO & $\mathbf{61.38{\pm}0.89}$ & $\mathbf{83.33{\pm}3.30}$ & $\mathbf{83.70{\pm}0.50}$ & $79.43{\pm}0.95$ & $\mathbf{76.96{\pm}0.54}$ \\
\bottomrule
\end{tabular}
\end{table}

\paragraph{ExPPO leads both ID and OOD coverage.}
ExPPO leads both coverage means and six task means, including a tie with F-GRPO on AIME25 (\cref{tab:nearest}). After training, ExPPO's OOD coverage exceeds initialization, while other methods generally show OOD forgetting.
%Its ID/OOD response accuracy is 27.98/59.98 versus 26.20/59.82 for F-GRPO (\cref{tab:nearest-accuracy}), complementing the coverage comparison.

\paragraph{Phi-3.5-mini.}
Across three seeds per method, ExPPO leads ID coverage and accuracy; REPO-R closely matches OOD coverage with less variation. GRPO-UR leads OOD accuracy but has lower, more variable coverage. F-GRPO has the lowest suite means and greatest coverage variation (\cref{tab:phi-fgrpo-main}).

\begin{table}[!htb]
\caption{Phi-3.5-mini at 300 steps: mean $\pm$ sample standard deviation across three independent training seeds per method. Each score is an unweighted four-task suite mean (\%). Coverage is the share of prompts with a verified success; response accuracy is the verified-response share. Bold: highest mean per column. Task and seed details: \cref{app:phi-fgrpo-ablation}.}
\label{tab:phi-fgrpo-main}
\centering
\small
\setlength{\tabcolsep}{4pt}
\begin{tabular*}{\linewidth}{@{\extracolsep{\fill}}lrrrr@{}}
\toprule
& \multicolumn{2}{c}{Coverage} & \multicolumn{2}{c}{Response accuracy} \\
\cmidrule(lr){2-3}\cmidrule(l){4-5}
Method & ID & OOD & ID & OOD \\
\midrule
F-GRPO & $13.88{\pm}5.76$ & $42.91{\pm}25.37$ & $6.42{\pm}0.81$ & $36.57{\pm}22.19$ \\
GRPO-UR & $21.66{\pm}0.75$ & $62.76{\pm}5.48$ & $9.79{\pm}1.11$ & $\mathbf{52.85{\pm}3.62}$ \\
REPO-R & $21.94{\pm}1.89$ & $\mathbf{66.35{\pm}0.38}$ & $9.64{\pm}0.19$ & $52.16{\pm}0.15$ \\
ExPPO & $\mathbf{25.67{\pm}1.31}$ & $66.33{\pm}1.59$ & $\mathbf{10.59{\pm}0.27}$ & $52.26{\pm}1.17$ \\
\bottomrule
\end{tabular*}
\end{table}

\FloatBarrier
\begin{figure}[H]
  \centering
  \setlength{\abovecaptionskip}{4pt}
  \includegraphics[width=\textwidth]{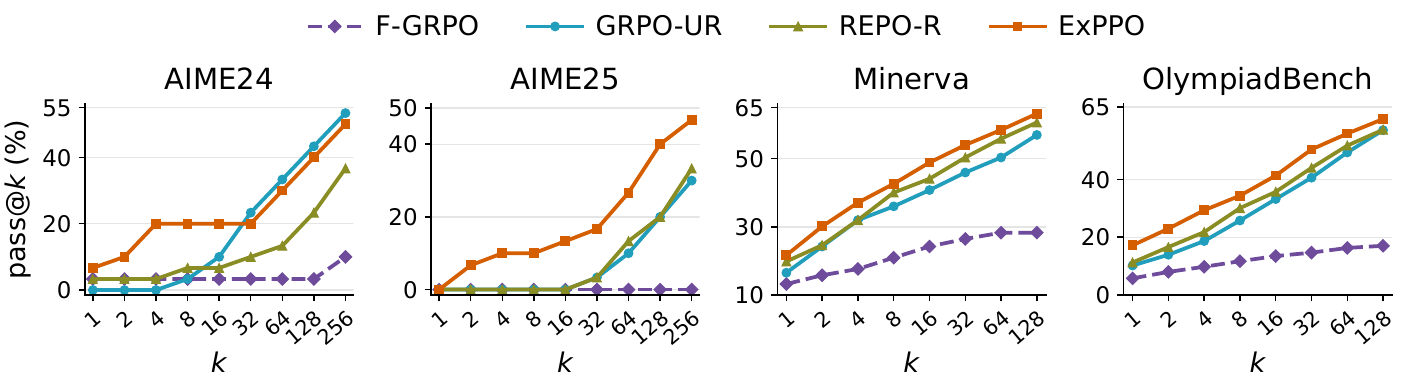}
  \caption{Phi-3.5-mini coverage (\%) from the first $k$ responses per prompt, for the methods in \cref{tab:phi-fgrpo-main}. ExPPO leads on three of four tasks at their largest evaluated budgets ($k=256$ for AIME24/25; $k=128$ otherwise), while GRPO-UR leads on AIME24. At the shared budget $k=128$, ExPPO's four-task macro coverage of 51.03 exceeds GRPO-UR's 44.34, REPO-R's 40.30, and F-GRPO's 12.17.}
  \label{fig:phi35-closest-high-k}
\end{figure}

\subsection{Beyond Mathematics: Code Reasoning}
\label{sec:code-reasoning}

Training Phi-3.5-mini on CodeReasoner-derived Python execution reasoning \citep{tang2025codereasoner} tests a second adaptation domain. ExPPO leads both code metrics and OOD means (\cref{tab:code-main}); its OOD response accuracy is 8.92 points above GRPO-H. This extends the accuracy--coverage benefit beyond mathematical training. See \cref{app:code-reasoning}.

\begin{table}[!htb]
\caption{Code adaptation and OOD generalization at the last checkpoint. Scores (\%): mean $\pm$ sample standard deviation across three independent Phi-3.5-mini training runs. Code: CodeReasoner; OOD: unweighted mean over GPQA, MMLU-Pro, Safety, and TruthfulQA-MC ($k=4$). Initialization: fixed base model. Bold: best trained mean.}
\label{tab:code-main}
% Source: experiment_evidence/phi35_code_reasoner_last_checkpoint.csv in the
% original upload. Code and OOD means were moved from the former Table 13.
% AUTHORS: Confirm the CodeReasoner coverage sampling budget k before submission;
% the supplied manuscript and summary CSV specify k=4 only for the OOD tasks.
\centering
\small
\setlength{\tabcolsep}{4pt}
\begin{tabular*}{\linewidth}{@{\extracolsep{\fill}}lrrrr@{}}
\toprule
& \multicolumn{2}{c}{Coverage} & \multicolumn{2}{c}{Response accuracy} \\
\cmidrule(lr){2-3}\cmidrule(l){4-5}
Method & Code & OOD & Code & OOD \\
\midrule
Initialization & 32.11 & 59.40 & 6.92 & 41.93 \\
GRPO & $85.94{\pm}0.00$ & $52.24{\pm}0.44$ & $75.22{\pm}0.04$ & $36.42{\pm}1.18$ \\
GRPO-H & $86.46{\pm}0.08$ & $60.51{\pm}1.20$ & $74.62{\pm}0.02$ & $42.10{\pm}0.34$ \\
ExPPO & $\mathbf{86.69{\pm}0.33}$ & $\mathbf{65.74{\pm}0.71}$ & $\mathbf{75.46{\pm}0.05}$ & $\mathbf{51.02{\pm}0.87}$ \\
\bottomrule
\end{tabular*}
\end{table}

\FloatBarrier
\section{Conclusion}
\label{sec:conclusion}

ExPPO combines bounded prompt-relative surprisal, pass-rate gating, and shared advantage-mass normalization to support exploration through response-level credit allocation. Our analysis identifies local conditions for gains in entropy and correct-mode discovery and characterizes how the rule preserves verifier polarity. Experiments across three backbones show improved in-distribution coverage with competitive out-of-distribution generalization, while sustaining policy entropy and increasing diversity among verified-correct responses. Large-sample evaluations and code-reasoning results further support the benefits across sampling budgets and training domains. ExPPO integrates into GRPO and DAPO using existing rollout statistics with negligible computational overhead.

% Keep non-scientific disclosures with the references, after the main text.
\clearpage
\section*{Reproducibility Statement}

\Cref{app:experimental-details,app:multi-answer-details,app:implementation} document model variants, training/evaluation settings, the final-answer parser, verifier, and metrics. The generation script and archived evidence cover checkpoint metrics, entropy histories, maximum-budget pools, the controlled $K=128$ rollout bank, paired-bootstrap summaries, Qwen decoding logs, sweeps, independent Phi-3.5-mini training runs, and derived summaries. A code package accompanies this submission as supplementary material.

\section*{AI Use Statement}
An AI assistant supported editing, reported-data arithmetic checks, and LaTeX formatting.
% AUTHORS: Extend this statement to cover all AI use across the full project,
% and verify the final manuscript and submission-form disclosure before submission.

\clearpage
\appendix
\crefalias{section}{appendix}
\crefalias{subsection}{appendix}
\crefname{appendix}{Appendix}{Appendices}
\Crefname{appendix}{Appendix}{Appendices}
\raggedbottom

\startcontents[appendix]
\section*{Appendix Contents}
\printcontents[appendix]{}{1}{\setcounter{tocdepth}{2}}
\clearpage

\section{Related Work}
\label{app:related}

\paragraph{RLVR objectives and support concentration.}
GRPO introduced critic-free group-relative optimization for mathematical reasoning, DAPO developed a stronger large-scale training recipe, and GSPO aligned importance weighting and clipping with sequence-level rewards \citep{shao2024deepseekmath,yu2025dapo,zheng2025gspo}. Subsequent studies report higher pass@1 together with lower large-$K$ coverage, narrower answer support, and stronger dependence on behaviors already present in the base policy \citep{yue2025rlcapacity,wu2025invisible}. MaxRL reweights low-success prompts through objectives that approach maximum likelihood as rollout compute grows \citep{tajwar2026maxrl}. \method{} examines the within-prompt allocation of credit across sampled responses.

\paragraph{Entropy control and set-level exploration.}
Entropy-focused work characterizes policy-entropy dynamics and develops explicit control through covariance clipping, adaptive regularization, advantage shaping, and asymmetric clipping \citep{cui2025entropy,zhang2025aer,petrenko2026entropy,zhang2026polarity,cheng2026gepo}. Set-level methods optimize the collective utility of several responses, including Pass@K Policy Optimization, analytical pass@$k$ training, and Poly-EPO \citep{walder2025pkpo,chen2025passk,orney2026polye}. Differential Smoothing modifies correct-trajectory rewards to counteract distribution sharpening \citep{gai2025smoothing}. These approaches motivate exploration measurements beyond pass@1. \method{} operates at response granularity and approximately preserves prompt-group absolute sequence-advantage mass while reallocating it across samples.

\paragraph{Rarity-aware and correct-support credit.}
The closest methods directly alter credit for underrepresented responses. Rewarding the Unlikely increases pressure on low-probability correct trajectories, and F-GRPO downweights high-success prompts to protect rare correct modes under finite groups \citep{he2025unlikely,plyusov2026fgrpo}. UCPO encourages a more uniform distribution within the correct set, and Cue-GRPO redistributes positive advantages using correct-strategy partitions \citep{lochab2026ucpo,cao2026cuegrpo}. \method{} specifies a prompt-relative allocation through group-standardized response surprisal, a bounded transformation, pass-rate gating, and absolute-advantage-mass normalization. The controlled task measures correct-mode yield and conditional correct-pair diversity, and \cref{tab:nearest} compares ExPPO with F-GRPO, GRPO-UR, and REPO-R. REPO-R controls token-level rescaling through entropy feedback. ExPPO uses prompt-relative response standardization, pass-rate gating, bounded corrections, and a shared absolute-mass budget.

%\paragraph{Relation to REPO-R.}
%REPO-R shapes both advantage signs using token-level surprisal and an adaptive entropy controller \citep{petrenko2026entropy}. Before sign clipping, its practical rule in \cref{eq:repo-r-implementation} and \method{}'s raw rule can be written as
%\[
%\begin{aligned}
%\widehat A^{\mathrm R}_{i,t}
%    &=A_i+\zeta|A_i|(-\ell_{i,t}),\\
%\widehat A^{\mathrm E}_i
%    &=A_i+\lambda|A_i|g_i
%      \tanh\!\left(\frac{s_i-\mu_{s,q}}{\max(\sigma_{s,q},\varepsilon)}\right).
%\end{aligned}
%\]
%The shared $|A_i|$ factor makes their relationship explicit. REPO-R controls token-level rescaling through entropy feedback. \method{} controls the relative allocation across complete responses within each prompt: group centering sets the common/rare boundary, the pass-rate gate allocates correction strength, the bound controls each perturbation, and the shared normalization fixes the absolute-mass budget up to numerical stabilization. These operations jointly define \method{}'s response-level redistribution rule.
%
\section{Scope and Limitations}
\label{app:limitations}

%\paragraph{Evaluation scope.}
%The evaluation covers three backbones, matched 900-step reasoning runs, and a four-method Phi-3.5-mini comparison with three training runs per method at 300 steps. Larger models, broader benchmark suites, and longer schedules would further characterize scaling with model capacity and optimization time.
%
%\paragraph{Local diagnostics and semantic rarity.}
%The common-checkpoint continuation in \cref{app:continuation-audit} characterizes one additional update from a fixed ExPPO-trained policy. Both paired diversity intervals span zero, and the positive accuracy interval relative to GRPO-H is unadjusted. On the update prompts, ExPPO point estimates trail GRPO. The coefficient audit measures response-level advantage allocation. In a separate length- and format-adjusted analysis, the correlations between surprisal and inverse semantic-mode frequency are $0.165$ at the sample level and $0.173$ at the mode level, with both confidence intervals spanning zero. Positive raw utility covariance accompanies uncertainty in the residualized alignment diagnostics. The theoretical conditions in \cref{eq:matched-kl-derivative} identify the residual alignment associated with local utility gains. Larger prompt panels and multiple starting policies would further characterize the connection between response surprisal, semantic-mode frequency, and accumulated diversity changes.
\paragraph{Evaluation scope.}
The evaluation covers three backbones, matched 900-step reasoning runs, and a four-method Phi-3.5-mini comparison with three training runs per method at 300 steps under limited computational resources. Larger models, broader benchmark suites, and longer schedules would further characterize scaling with model capacity and optimization time.

\section{Binary-Reward GRPO and Mode-Level Credit}
\label{app:binary-grpo}

This derivation gives the closed-form exposure effect used in \cref{sec:motivation}. Consider a mixed group with binary rewards $r_i\in\{0,1\}$ and sampled pass rate $\rho_q\in(0,1)$. Ignoring the numerical constant in the denominator, the group reward mean and standard deviation are $\mu_{r,q}=\rho_q$ and $\sigma_{r,q}=\sqrt{\rho_q(1-\rho_q)}$. Therefore,
\begin{equation}
    A_i=
    \begin{cases}
        \sqrt{(1-\rho_q)/\rho_q}, & r_i=1,\\[2pt]
        -\sqrt{\rho_q/(1-\rho_q)}, & r_i=0.
    \end{cases}
    \label{eq:binary-grpo}
\end{equation}
All successes receive the same positive coefficient and all failures receive the same negative coefficient. If a correct mode $m$ appears $n_m$ times in the group, its aggregate positive coefficient is $C_m=n_m\sqrt{(1-\rho_q)/\rho_q}$. This identity is the mode-level multiplicity effect that motivates the controlled experiment.

\begin{proposition}[Multiplicity changes relative mode odds]
\label[proposition]{prop:multiplicity-logodds}
Let $\mathcal{M}_q$ be a finite set of response modes for prompt $q$, let $p_z(m\mid q)=e^{z_m}/\sum_{a\in\mathcal{M}_q}e^{z_a}$ be a categorical mode policy with logit $z_m$, and let $m_i$ be the mode containing response $y_i$. Define $C_m=\sum_{i:m_i=m}A_i$, the empirical policy-gradient surrogate $\mathcal{L}_q(z)=\sum_{i=1}^{G}A_i\log p_z(m_i\mid q)$, and the relative log-odds $\Delta_{mm'}(p)=\log p(m\mid q)-\log p(m'\mid q)$. Because group-relative advantages satisfy $\sum_iA_i=0$, one gradient-ascent step $z_m^{+}=z_m+\eta\,\partial\mathcal{L}_q/\partial z_m$ with step size $\eta>0$ gives, for any two correct modes $m$ and $m'$,
\begin{equation}
\Delta_{mm'}(p_{z^{+}})-\Delta_{mm'}(p_z)=\eta(C_m-C_{m'})=\eta A^{+}(n_m-n_{m'}).
\label{eq:multiplicity-logodds}
\end{equation}
A count advantage $n_m>n_{m'}$ therefore gives mode $m$ a larger relative log-odds update.
\end{proposition}

\paragraph{Proof.}
For the softmax mode policy, $\partial\mathcal{L}_q/\partial z_m=C_m-p_z(m\mid q)\sum_iA_i=C_m$. The identity $\Delta_{mm'}(p_z)=z_m-z_{m'}$ then gives \cref{eq:multiplicity-logodds}. This mode-level derivation follows the score-function policy-gradient update \citep{williams1992reinforce}.

\paragraph{GRPO and ExPPO in the same logit geometry.}
For arbitrary detached sample scores $h_i$, set $p_m=p_z(m\mid q)$, $H_m=\sum_{i:m_i=m}h_i$, and $H_\Sigma=\sum_i h_i$. Differentiating $\mathcal L_q^h=\sum_i h_i\log p_z(m_i\mid q)$ gives
\[
\frac{\partial\mathcal L_q^h}{\partial z_m}
=H_m-p_mH_\Sigma.
\]
Let $C_m^B=\sum_{i:m_i=m}B_i$ and $B_\Sigma=\sum_i B_i$. Applying the same gradient-ascent step size $\eta$ to $h_i=A_i$ and $h_i=\widetilde A_i$ yields the respective relative log-odds changes
\[
\begin{aligned}
\delta^A_{mm'}&=\eta(C_m-C_{m'}),\\
\delta^{\mathrm E}_{mm'}
&=\alpha_q\delta^A_{mm'}
 +\eta\alpha_q\lambda\big[C_m^B-C_{m'}^B-(p_m-p_{m'})B_\Sigma\big].
\end{aligned}
\]
The first term records shared rescaling; the bracket gives the additional allocation contribution in the same geometry as Proposition~\ref{prop:multiplicity-logodds}. For a rare/common pair, a positive bracket shifts relative odds toward the rare mode compared with the shared-scale base update. The $B_\Sigma$ term accounts for the group-sum change introduced by shaping.

The effective score $h_{\mathrm{eff}}(m)=H_m-p_mH_\Sigma$ also gives the exact updated distribution,
\[
p_{z^+}(m\mid q)
=\frac{p_z(m\mid q)e^{\eta h_{\mathrm{eff}}(m)}}
       {\sum_a p_z(a\mid q)e^{\eta h_{\mathrm{eff}}(a)}}.
\]
Thus the sampled logit step is an exponential tilt with its gradient-induced score. Setting $h=\widetilde A$ in Theorem~\ref{thm:rare-preference} evaluates the response-level advantage allocation within the same exponential-tilt family.

\paragraph{Neural gradients and induced response scores.}
At $\theta_{\mathrm{old}}$, the token importance ratios equal one and the clipping interval contains them in its interior. For the sampled policy-loss term, let $\omega_{i,t}$ denote the fixed token aggregation weights and evaluate the following derivatives at $\theta=\theta_{\mathrm{old}}$:
\[
f_i=\sum_t\omega_{i,t}\nabla_\theta\log\pi_\theta(y_{i,t}\mid x_{i,t}),
\qquad
\mathcal G_q(h)=\sum_i h_i f_i.
\]
The prompt's gradient contribution satisfies $\mathcal G_q(\widetilde A)=\alpha_q\mathcal G_q(A)+\alpha_q\lambda\mathcal G_q(B)$. Response averaging uses $\omega_{i,t}=1/(GT_i)$; token-mean aggregation uses the reciprocal of the total generated-token count in the loss aggregation batch. These expressions retain parameter sharing through the vectors $f_i$.
For a fixed parameter direction $d$ and differentiable policy on a finite response space, the score $h_d(y)=\nabla_\theta\log\pi_\theta(y\mid q)^\top d$, evaluated at $\theta=\theta_{\mathrm{old}}$, has $\E_{\pi_{\mathrm{old}}}[h_d]=0$ and satisfies
\[
\left.\frac{d\pi_{\theta_{\mathrm{old}}+\eta d}(y\mid q)}{d\eta}\right|_0
=\pi_{\mathrm{old}}(y\mid q)h_d(y)
=\left.\frac{d\pi_{\eta,h_d}(y\mid q)}{d\eta}\right|_0.
\]
This identity connects a chosen optimizer direction to the local density model in \cref{lem:tilt}.

\section{GRPO Background and Entropy Bonuses}
\label{app:grpo-details}

\subsection{GRPO and GRPO-H Objective Details}
\label{app:grpo-objectives}

\Cref{eq:grpo-objective,eq:entropy-bonus} give the objectives used in the main text. Here $\mathcal B$ denotes joint sampling of a prompt from $\mathcal D$ and a group of $G$ responses from $\pi_{\mathrm{old}}(\cdot\mid q)$. For realized response $y_i$, the sequence probability factors as
$\pi_{\mathrm{old}}(y_i\mid q)=\prod_{t=1}^{T_i}\pi_{\mathrm{old}}(y_{i,t}\mid x_{i,t})$.
GRPO uses the detached advantage in \cref{eq:grpo-advantage} and the token importance ratio
\[
w_{i,t}(\theta)=\frac{\pi_\theta(y_{i,t}\mid x_{i,t})}{\pi_{\mathrm{old}}(y_{i,t}\mid x_{i,t})}.
\]
For clipping radius $\epsilon_{\mathrm{clip}}>0$, the clipped surrogate is
\[
\ell_{\mathrm{clip}}(w,A)=\min\{wA,\operatorname{clip}(w,1-\epsilon_{\mathrm{clip}},1+\epsilon_{\mathrm{clip}})A\}.
\]
The reference-policy penalty is
$d_{i,t}(\theta)=D_{\mathrm{KL}}(\pi_\theta(\cdot\mid x_{i,t})\|\pi_{\mathrm{ref}}(\cdot\mid x_{i,t}))$,
where $\pi_{\mathrm{ref}}$ is fixed and $\beta_{\mathrm{KL}}\geq0$ controls its strength.
Canonical response averaging uses
$\langle z_{i,t}\rangle=G^{-1}\sum_{i=1}^{G}T_i^{-1}\sum_{t=1}^{T_i}z_{i,t}$ \citep{shao2024deepseekmath}.
Our implementation uses token-mean aggregation, as specified in \cref{app:experimental-details}; the response-level shaping applies with either convention.

The \grpoh{} objective adds Shannon next-token entropy,
$H(\pi_\theta(\cdot\mid x))=-\sum_a\pi_\theta(a\mid x)\log\pi_\theta(a\mid x)$,
with coefficient $\beta_{\mathrm{ent}}\geq0$ \citep{williams1991entropy,mnih2016a3c,schulman2017ppo}.
This term averages uncertainty over sampled token contexts using the same aggregation as the policy loss.

\subsection{Entropy Bonuses in MuJoCo and Their Connection to RLVR}
\label{app:entropy-control}

\paragraph{Evidence from continuous control.}
In a large PPO study on five MuJoCo tasks, \citet{andrychowicz2021matters} find similar performance across most tested regularizers, with entropy constraints benefiting HalfCheetah. They suggest that careful initialization already supplies useful exploration and that PPO clipping controls policy changes. \citet{ahmed2019entropy} show that higher entropy can smooth policy optimization in some environments. Together, these results tie the value of an entropy bonus to the initial policy and task dynamics.

\paragraph{Action spread and task direction.}
For a diagonal Gaussian policy at state $x$, the differential entropy is
\begin{equation}
H\!\left(\mathcal N(\mu_\theta(x),\operatorname{diag}\sigma_\theta^2(x))\right)
=\frac{d_a}{2}\log(2\pi e)+\sum_{j=1}^{d_a}\log\sigma_{\theta,j}(x),
\label{eq:gaussian-entropy}
\end{equation}
where $d_a$ is the action dimension. With separate mean and scale parameters, the direct entropy gradient changes action spread and has zero derivative with respect to the mean; the reward advantage supplies the task direction. Increasing spread can therefore spend samples on low-return actions. Its balance with reward depends on the bonus coefficient, reward scale, and initialization. This formula describes the Gaussian before action clipping or squashing.

\paragraph{Local entropy and long-horizon credit.}
A direct entropy term at fixed rollout states promotes uncertainty at those states. A maximum-entropy return also credits earlier actions for reaching states with high future entropy \citep{haarnoja2018sac}. \citet{choe2024eapo} separate reward and entropy advantage estimates and match or improve entropy-bonus PPO/TRPO on their MuJoCo comparisons, whose main experiments use discretized actions. These results show how exploration benefits from assigning credit across a trajectory.

\paragraph{GRPO-H and verifier credit.}
The bonus in \cref{eq:entropy-bonus} supplies a reward-independent gradient at each sampled prefix. For a softmax token policy $p_a=\exp(z_a)/\sum_b\exp(z_b)$,
\begin{equation}
\frac{\partial H(p)}{\partial z_a}=p_a\bigl[-\log p_a-H(p)\bigr].
\label{eq:direct-entropy-gradient}
\end{equation}
Thus the bonus favors tokens with above-average surprisal, including continuations that fail the verifier, while the GRPO term retains equal sequence advantages for equal rewards. Its scale also matters: \citet{cui2025entropy} report weak entropy control at small coefficients, rapid entropy growth at a large coefficient, and performance at or below the compared baselines at a coefficient that stabilizes entropy. These mechanisms help explain the high entropy and low ID coverage of \grpoh{} in our mathematical runs (\cref{fig:entropy-dynamics,fig:training-macro}). The code experiment shows gains from \grpoh{} in a different task regime (\cref{tab:code-main}). \method{} ties exploration credit to the verifier advantage and prompt pass rate, with bounded shaping and shared normalization; \cref{eq:entropy-decomposition} explains why diversity among correct responses is the relevant target.

\section{Experimental Details}
\label{app:experimental-details}

\paragraph{Two instruction checkpoints and one base checkpoint span model regimes.}
The short names in the main paper correspond to \path{Qwen/Qwen2.5-7B-Instruct} (\textbf{Qwen2.5-7B}), \path{Qwen/Qwen3-0.6B} (\textbf{Qwen3-0.6B}), and \path{microsoft/Phi-3.5-mini-instruct} (\textbf{Phi-3.5-mini}). Qwen2.5 and Phi use their instruction templates, while Qwen3 uses the base checkpoint and tokenizer. Every method for a model starts from the same checkpoint and tokenizer.

\paragraph{Shared optimization isolates advantage shaping.}
Training uses the fixed 17,398-prompt DAPO-Math-17k set, with left truncation and removal of prompts longer than 2,048 tokens. The maximum response length is 10,240 tokens. Each optimization step samples $G=16$ responses per prompt with temperature 1.0, top-$p=1.0$, and top-$k$ disabled. The prompt batch size is 64, the policy mini-batch size is 32, and the rollout-generation batch size is 32. We optimize with AdamW at learning rate $10^{-6}$, 10 warmup steps, weight decay 0.1, and gradient clipping at 1.0. The policy loss uses token-mean aggregation, symmetric PPO clipping $(\epsilon_{\mathrm{low}},\epsilon_{\mathrm{high}})=(0.2,0.2)$, zero reward KL coefficient, and zero actor KL coefficient. The main reasoning-benchmark runs use 900 steps; the Phi-3.5-mini coefficient, component, and four-method baseline comparisons use checkpoint 300. Core methods save every 100 steps; GRPO-UR, REPO-R, and F-GRPO save every 200 steps and at the final step.

GRPO uses the group-standardized verifier advantage and a zero auxiliary exploration coefficient. ExPPO uses \cref{eq:exppo-shaping} with $\lambda=0.2$. GRPO-H uses \cref{eq:entropy-bonus} with $\beta_{\mathrm{ent}}=0.001$. In the closest-method experiment, GRPO-UR uses unlikeliness coefficient 0.25, REPO-R uses $(\zeta_{\mathrm{init}},\zeta_{\min},\zeta_{\max})=(10^{-3},10^{-4},0.05)$, and F-GRPO uses focal exponent $\gamma=0.5$. These modifiers are mutually exclusive and otherwise retain the GRPO configuration.

\paragraph{Closest baselines target rarity and entropy through distinct signals.}
F-GRPO multiplies every normalized advantage in prompt group $q$ by $(1-\rho_q)^\gamma$, with $\gamma=0.5$ \citep{plyusov2026fgrpo}. Our GRPO-UR reproduction follows Rewarding the Unlikely \citep{he2025unlikely}. Let $\bar\ell_i=T_i^{-1}\sum_t\log\pi_{\mathrm{old}}(y_{i,t}\mid x_{i,t})$ and let $\mathrm{rank}_i=\operatorname{rank}_{\uparrow}(\bar\ell_i)\in\{0,\ldots,G-1\}$, where rank zero is least likely. Before group normalization, nonzero-signal samples use
\begin{equation}
    \widehat r_i^{\mathrm{UR}}=r_i\left(1-\beta_{\mathrm{UR}}\frac{\mathrm{rank}_i}{G}\right),
    \qquad \beta_{\mathrm{UR}}=0.25.
    \label{eq:grpo-ur-implementation}
\end{equation}
All-pass and all-fail groups retain their original zero advantage. Our REPO-R reproduction \citep{petrenko2026entropy} applies the detached current-policy token log probability $\ell_{i,t}=\operatorname{sg}[\log\pi_\theta(y_{i,t}\mid x_{i,t})]$ to the GRPO advantage:
\begin{equation}
\widetilde A^{\mathrm{R}}_{i,t}=
\begin{cases}
\max\{A_i(1-\zeta\ell_{i,t}),0\}, & A_i>0,\\
\min\{A_i(1+\zeta\ell_{i,t}),0\}, & A_i<0,\\
0, & A_i=0.
\end{cases}
\label{eq:repo-r-implementation}
\end{equation}
Its target is the token-mean entropy observed on the first step. The signed controller doubles or halves $|\zeta|$ according to the target-entropy error and bounds its nonzero magnitude to $[10^{-4},0.05]$, starting from $10^{-3}$.

\paragraph{Fixed budgets separate coverage from response accuracy.}
Standard evaluations sample with temperature 0.7, top-$p=0.95$, and at most 10,240 new tokens. Generation ends at the tokenizer end-of-sequence token or the length cap. We render each model's serialized reasoning prompt with its tokenizer and request a boxed final answer; multiple-choice records additionally request one option letter. The benchmark-specific rule-based extractor scores the final answer.

AIME24 and AIME25 use $k=16$, Minerva uses $k=8$, and OlympiadBench, GPQA, MMLU-Pro, Safety, and TruthfulQA-MC use $k=4$. For prompt $q$, avg@$k$ is the fraction of its $k$ responses that pass the verifier; reported benchmark scores average this quantity over prompts and call it \emph{response accuracy}. The main-paper coverage measure follows \cref{sec:exp-setup}. Checkpoint trends use one fixed decoding configuration. The Qwen2.5-7B closest-method comparison evaluates one fixed step-900 checkpoint per method with three decoding seeds. Its sample standard deviations quantify decoding variability conditional on the trained checkpoint; suite averages are computed within each decoding seed before aggregation.

\begin{table}[H]
\caption{Reasoning-evaluation suites and prompt counts. Benchmark: evaluation set; Suite: In-Domain (ID) adaptation or Out-of-Domain (OOD) generalization; Prompts: evaluated questions. Each suite contains four tasks, and suite averages are unweighted.}
\label{tab:eval-files}
\centering
\scriptsize
\setlength{\tabcolsep}{3.5pt}
\begin{tabular}{lrr}
\toprule
Benchmark & Suite & Prompts \\
\midrule
AIME24 & ID & 30 \\
AIME25 & ID & 30 \\
Minerva & ID & 272 \\
OlympiadBench & ID & 675 \\
GPQA & OOD & 448 \\
MMLU-Pro & OOD & 70 \\
Safety & OOD & 1,000 \\
TruthfulQA-MC & OOD & 684 \\
\bottomrule
\end{tabular}
\end{table}

% AUTHORS: Confirm the source repository, configuration/split, revision, and
% conversion script for each ID benchmark before submission.
The ID evaluation uses the four benchmark files summarized in \cref{tab:eval-files}. The OOD files use GPQA's \texttt{gpqa\_main}/train split, MMLU-Pro validation, the train split of the pairwise safety set\footnote{\url{https://huggingface.co/datasets/ThWu/safety_benchmark}}, and TruthfulQA MC1. TruthfulQA-MC retains questions with at least four choices and deterministically samples four choices when more are available. This ID/OOD setup adapts the checkpoint-wise evaluation of \citet{jin2025oodforgetting,jin2026rpsft}.

At each displayed $k$, large-sample coverage is the percentage of prompts with at least one verified success among their first $k$ generated responses. This ordered-prefix estimate uses the responses available at each displayed budget. The maximum displayed budget is 256 for AIME24/25 and 128 for Minerva/OlympiadBench. The entropy curves read \texttt{actor/entropy} against \texttt{training/global\_step}; duplicate resumed-log rows are deduplicated by step, and all plots stop at the final checkpoint.

\FloatBarrier
\subsection{Measured Training Cost}
\label{app:overhead}

\begin{table}[H]
\caption{Qwen2.5-7B-Instruct training cost on one eight-H200 node. Although the configured training prompt batch is 64, the synchronous data loader uses a generation batch of 32 prompts with 16 responses each, yielding 512 responses per logged step; Samples/s therefore equals 512 times Steps/s. Each profile runs for 100 steps, and throughput and per-GPU peak memory use steps 11--100. Estimated GPU-hours scale the full eight-GPU allocation wall time of the corresponding 100-step job by nine; $\dagger$ marks the separately profiled F-GRPO job. GRPO-H uses the unchunked entropy-loss profile with activation checkpointing disabled. Peak alloc./reserved are GiB per GPU; GRPO-UR abbreviates Rewarding the Unlikely.}
\label{tab:overhead}
\centering
\small
\setlength{\tabcolsep}{4.0pt}
\resizebox{\linewidth}{!}{%
\begin{tabular}{lrrrrr}
\toprule
Method & GPU-hours & Samples/s & Steps/s & Peak alloc. & Peak reserved \\
\midrule
GRPO & 151.70 & 6.997 & 0.01367 & 88.9 GiB & 101.2 GiB \\
GRPO-H & 157.76 & 6.710 & 0.01311 & 88.5 GiB & 100.8 GiB \\
GRPO-UR & 136.18 & 8.107 & 0.01583 & 86.5 GiB & 98.6 GiB \\
REPO-R & 151.66 & 7.055 & 0.01378 & 87.8 GiB & 100.0 GiB \\
F-GRPO & 146.08 & 7.289 & 0.01424 & 87.3 GiB & 99.6 GiB \\
ExPPO & 150.12 & 7.230 & 0.01412 & 87.7 GiB & 99.9 GiB \\
\bottomrule
\end{tabular}}
\end{table}

\paragraph{ExPPO matches GRPO's compute envelope.}
ExPPO and GRPO occupy the same end-to-end compute band under the common schedule. GRPO-UR and F-GRPO are faster, GRPO-H takes more time, and all six methods remain closely matched in peak memory.

\section{Optimization Implementation}
\label{app:implementation}

\paragraph{ExPPO transfers across GRPO and DAPO objectives.}
The GRPO implementation evaluates \cref{eq:grpo-objective} with the substitution $A_i\leftarrow\widetilde A_i$. The DAPO implementation applies the same sequence-level substitution inside its configured asymmetric clipped objective and retains its dynamic sampling, token-level aggregation, and overlong-response shaping. This construction keeps the rollout and optimizer pipeline fixed while changing the allocation of detached response advantages.

\paragraph{GRPO-H supplies a policy-wide exploration contrast.}
\Cref{eq:entropy-bonus} defines the \grpoh{} objective used in the main comparison. The implementation computes token-distribution entropy over the same generated-token mask and applies the same token aggregation convention as the policy loss. This baseline supplies a policy-wide exploration signal for comparison with the response-level allocation implemented by \method{}.

\paragraph{ExPPO adds one bounded detached correction per response.} \Cref{alg:exppo} summarizes the response-level shaping for one prompt group.
\begin{algorithm}[ht]
\caption{Exploration-Preserving Policy Optimization for one prompt group}
\label{alg:exppo}
\begin{algorithmic}[1]
\Require Prompt $q$, rollout policy $\pi_{\mathrm{old}}$, group size $G$, coefficient $\lambda\in[0,1]$
\State Sample responses $\{y_i\}_{i=1}^{G}$ and compute rewards $\{r_i\}_{i=1}^{G}$
\State Compute detached base advantages $A_i$ and verifier-defined success indicators $c_i$
\State Compute $s_i$ and $u_i$ using \cref{eq:rarity}; set $\rho_q\gets G^{-1}\sum_i c_i$
\State Compute gates $g_i$ and raw shaped advantages $\widehat A_i$ using \cref{eq:exppo-shaping}
\State Normalize $\widehat A_i$ into $\widetilde A_i$ using \cref{eq:normalization}
\State Broadcast $\widetilde A_i$ across response tokens and optimize the selected base policy loss
\end{algorithmic}
\end{algorithm}

\section{Notation and Probability Conventions}
\label{app:notation}

\paragraph{Random responses and realized values.}
For a fixed prompt $q$, $Y_i\sim\pi_{\mathrm{old}}(\cdot\mid q)$ is a random rollout and $y_i$ its realized value. After observing $y_{1:G}$, indexed quantities such as $T_i$, $A_i$, and $B_i$, and group statistics $\rho_q$ and $\alpha_q$, are fixed. Uppercase $A_i$ follows standard advantage notation. A fresh response $Y\sim\pi_{\mathrm{old}}(\cdot\mid q)$ induces the random variables $A=A(Y)$, $B=B(Y)$, $S=S(Y)$, and $V=V(Y)$; $A(y)$, $B(y)$, $S(y)$, and $V(y)$ are their values at response $y$. The verifier $R(q,\cdot)$ is a deterministic function; under binary rewards, $C=R(q,Y)$ is the random success indicator. The prompt success probability $p_q$ is a population quantity, and $\rho_q$ is the observed group pass rate.

\paragraph{Frozen statistics and probability measures.}
Local analysis fixes $q$, $y_{1:G}$, and all group statistics before varying the response distribution. It uses the finite response space induced by the generation length limit. For any response in the rollout support, set $A(y)=(R(q,y)-\mu_{r,q})/(\sigma_{r,q}+\varepsilon)$ and $s(y)=-T(y)^{-1}\log\pi_{\mathrm{old}}(y\mid q)$; compute $u(y)$, $g(y)$, $B(y)$, and $\widetilde A(y)$ using the same frozen group statistics as in the method. At $y=y_i$, these functions equal their indexed values. Unless a measure is shown explicitly, expectations, variances, and covariances in the proofs use $Y\sim\pi_0=\pi_{\mathrm{old}}(\cdot\mid q)$. Empirical group covariance uses the uniform measure on the realized sample indices:
\[
\widehat{\Cov}_q(v,z)=\frac1G\sum_{i=1}^{G}(v_i-\bar v_q)(z_i-\bar z_q),
\qquad \bar v_q=\frac1G\sum_{i=1}^{G}v_i,\quad \bar z_q=\frac1G\sum_{i=1}^{G}z_i.
\]
Thus \cref{prop:rarity-alignment} concerns observed, length-normalized surprisal, and \cref{eq:entropy-slope} concerns population full-response surprisal. The scalar $\alpha_q$ remains fixed as the tilt step $\eta$ varies. It may vary with $\lambda$; positive shared scaling cancels from $\Gamma_V$.

\begingroup
\small
\setlength{\tabcolsep}{5pt}
\renewcommand{\arraystretch}{1.0}
\begin{longtable}{@{}>{\raggedright\arraybackslash}p{0.29\linewidth}>{\raggedright\arraybackslash}p{0.675\linewidth}@{}}
\caption{Notation for the method and analysis. Indexed quantities are evaluated on the realized rollout group.}\label{tab:notation}\\
\toprule
Symbol & Meaning \\
\midrule
\endfirsthead
\multicolumn{2}{@{}l}{\tablename~\thetable\ (continued)}\\
\toprule
Symbol & Meaning \\
\midrule
\endhead
\bottomrule
\endfoot
\multicolumn{2}{@{}l}{\textbf{Rollouts, objectives, and shaping}}\\*
$q,\mathcal D,G$ & Prompt, training prompt distribution, and responses per prompt group. \\
$Y_i,y_i;\ Y,y$ & Random rollout and realized rollout; fresh random response and a value in its support. \\
$T_i,y_{i,t},x_{i,t}$ & Realized response length, token $t$, and context $(q,y_{i,<t})$. \\
$\theta,\pi_\theta,\pi_{\mathrm{old}},\pi_{\mathrm{ref}}$ & Parameters; current, frozen rollout, and fixed reference policies. \\
$R(q,y),r_i;\ C,c_i$ & Verifier and realized reward; random and realized success indicators ($c_i=r_i$ for binary rewards). \\
$\mu_{r,q},\sigma_{r,q},\varepsilon$ & Group reward mean, reward standard deviation, and numerical stabilizer. \\
$A_i,A(y),A=A(Y)$ & Realized advantage, its frozen-statistic response function, and the induced random variable. \\
$\mathcal B,\langle\cdot\rangle$ & Joint prompt/group sampling and canonical response-averaged token aggregation. \\
$w_{i,t},\ell_{\mathrm{clip}},\epsilon_{\mathrm{clip}}$ & Token importance ratio, clipped surrogate, and clipping radius. \\
$d_{i,t},\beta_{\mathrm{KL}},\beta_{\mathrm{ent}}$ & Reference-policy KL divergence and the reference-KL and entropy coefficients. \\
$d_a,\mu_\theta(x),\sigma_\theta(x)$ & Action dimension, Gaussian policy mean, and componentwise standard deviation in \cref{eq:gaussian-entropy}. \\
$z_a,p_a$ & Token logit and softmax probability at a fixed prefix in \cref{eq:direct-entropy-gradient}. \\
$s_i,\mu_{s,q},\sigma_{s,q}$ & Realized length-normalized response surprisal and its group mean and standard deviation. \\
$u_i,\rho_q,g_i,\lambda$ & Bounded relative surprisal, observed pass rate, pass-rate gate, and shaping strength. \\
$B_i,\widehat A_i,\widetilde A_i$ & Correction $|A_i|g_i u_i$, raw advantage $A_i+\lambda B_i$, and normalized advantage. \\
$\alpha_q,M_i$ & Shared absolute-mass scale and raw multiplier $1+\lambda g_i\sgn(A_i)u_i$. \\
$\psi_\rho(u)$ & Gated score $u((1-\rho)\ind[u>0]+\rho\ind[u<0])$. \\
$\bar M_A,\xi_q,\kappa(\lambda)$ & Absolute-advantage-weighted mean multiplier, stabilizer divided by base absolute mass, and relative multiplier-spread bound. \\
\addlinespace[3pt]
\multicolumn{2}{@{}l}{\textbf{Local response updates and discovery}}\\*
$\pi_0,\pi_{\eta,h},\eta$ & Baseline response law, exponential tilt with score $h$, and tilt step size. \\
$h(q,y),h_d(y)$ & Frozen response score and score induced by parameter direction $d$. \\
$S(y),S=S(Y)$ & Full-response surprisal $-\log\pi_0(y)$ and its random-variable form; $S(y_i)=T_i s_i$. \\
$H,H_b,D_{\mathrm{KL}}$ & Shannon entropy, binary entropy, and KL divergence. \\
$\widehat{\Cov}_q,\Cov,\Var$ & Uniform empirical group covariance, population covariance, and population variance. \\
$V(y),J_V(\pi)$ & Frozen response utility and its expectation under $\pi$. \\
$\delta,\eta_h(\delta),\Gamma_V(h)$ & KL budget, score-specific step attaining it, and $\Cov(V,h)/\sqrt{\Var(h)}$. \\
$B_\perp$ & Correction after removing its covariance-aligned component along $A$. \\
$\mathcal M_q,\mathcal M_q^+,m(y)$ & Response-mode partition, its correct modes, and the mode of response $y$. \\
$p_m(\pi),p_m,p_q,p_+$ & Mode mass under $\pi$; baseline mode mass; success mass; the proof notation $p_+=\sum_{m\in\mathcal M_q^+}p_m=p_q$. \\
$K,F_K,V_K$ & Evaluation sample budget, expected number of distinct correct modes, and its frozen local utility. \\
$w_m,Q,D_{\mathrm{pair}},V_D$ & Correct-conditional mode mass, concentration $Q=\sum_m w_m^2$, pair disagreement $1-Q$, and its frozen local utility. \\
$\pi_\tau,\dot p_m,\dot p_+$ & Differentiable policy path and mode/success-mass derivatives at $\tau=0$. \\
$Z_h(\eta),K_h(\eta),\nu$ & Tilt normalizer, its logarithm, and the normalization Lagrange multiplier. \\
\addlinespace[3pt]
\multicolumn{2}{@{}l}{\textbf{Mode logits, gradients, and finite-$\lambda$ conditions}}\\*
$z_m,p_z,m_i,n_m,A^+,A^-$ & Mode logit, softmax mode policy, sampled mode, its count, and positive/negative binary-GRPO advantages. \\
$C_m,C_m^B,B_\Sigma$ & Mode-summed base credit, mode-summed correction, and total group correction. \\
$\mathcal L_q^h,H_m,H_\Sigma,h_{\mathrm{eff}}$ & Sampled logit surrogate, mode/group score sums, and effective logit-step score. \\
$\Delta_{ij},\Delta_{mm'},\delta^A_{mm'},\delta^{\mathrm E}_{mm'}$ & Response/mode relative log-odds and their base/ExPPO logit-step changes. \\
$\omega_{i,t},f_i,\mathcal G_q(h),d$ & Token aggregation weight, response gradient feature, prompt gradient, and parameter update direction. \\
$a,b,v,c,d$ & Scalar proof abbreviations: $\Cov(V,A)$, $\Cov(V,B)$, $\Var(A)$, $\Cov(A,B)$, and $\Var(B)$. Here $d$ is a scalar variance. \\
$a_R,b_R,b_D$ & Binary-advantage coefficients in $A=a_R C+b_R$ and diversity alignment $b_D=\Cov(V_D,B)$. \\
$P_0,P_1,\lambda_{\mathrm{num}},\lambda_{\mathrm{perf}}$ & Coefficients of the finite-gain inequality and its numerator/utility thresholds. \\
$C_B^H,\lambda_H$ & Entropy alignment $\Cov(S,B)$ and the nonnegative-entropy-slope threshold. \\
\end{longtable}
\endgroup

\section{Proofs}
\label{app:proofs}

\paragraph{The local-update lemma and local identities.}
For fixed $q$, abbreviate $\pi_0(y)=\pi_{\mathrm{old}}(y\mid q)$ and $h(y)=h(q,y)$. In expectations and covariances, $h$, $V$, and $S$ mean $h(Y)$, $V(Y)$, and $S(Y)$ under the indicated response law. Adding a Lagrange multiplier $\nu$ for $\sum_y\pi(y)=1$ to \cref{eq:local-kl} gives
\begin{equation}
    \mathcal{J}(\pi,\nu)=
    \sum_y\pi(y)h(y)
    -\frac{1}{\eta}\sum_y\pi(y)\log\frac{\pi(y)}{\pi_0(y)}
    +\nu\left(\sum_y\pi(y)-1\right).
\end{equation}
Stationarity with respect to $\pi(y)$ implies $\log(\pi(y)/\pi_0(y))=\eta h(y)+\eta\nu-1$. The last two terms are constant in $y$, and normalization yields \cref{eq:tilt}.

Let $Z_h(\eta)=\E_{\pi_0}[e^{\eta h}]$. Differentiating the normalized tilt gives
\begin{equation}
    \frac{d\pi_{\eta,h}(y)}{d\eta}
    =\pi_{\eta,h}(y)
    \left(h(y)-\E_{\pi_{\eta,h}}[h]\right).
    \label{eq:tilt-derivative}
\end{equation}
At $\eta=0$, $\pi_{\eta,h}=\pi_0$, so for any fixed $V$,
\begin{equation}
    \left.\frac{d}{d\eta}\E_{\pi_{\eta,h}}[V]\right|_0
    =\E_{\pi_0}\!\left[V(h-\E[h])\right]
    =\Cov_{\pi_0}(V,h).
\end{equation}

For response entropy $H(\pi_{\eta,h})=-\sum_y\pi_{\eta,h}(y)\log\pi_{\eta,h}(y)$, the derivative of the normalization term sums to zero. Defining $S(y)=-\log\pi_0(y)$ and substituting \cref{eq:tilt-derivative} at zero yields
\begin{equation}
    \left.\frac{dH(\pi_{\eta,h})}{d\eta}\right|_0
    =\Cov_{\pi_0}(S,h),
    \label{eq:entropy-proof}
\end{equation}
which proves \cref{eq:entropy-slope} after setting $h=\alpha_q(A+\lambda B)$.

Finally, writing $K_h(\eta)=\log Z_h(\eta)$ gives
\begin{equation}
    D_{\mathrm{KL}}(\pi_{\eta,h}\|\pi_0)
    =\eta K_h'(\eta)-K_h(\eta)
    =\frac{\eta^2}{2}\Var_{\pi_0}(h)+O(\eta^3).
    \label{eq:kl-expansion}
\end{equation}
Solving for the score-specific step size $\eta_h(\delta)$ at a small KL budget $\delta$ and substituting into the utility expansion gives
\begin{equation}
    J_V(\pi_{\eta_h,h})-J_V(\pi_0)
    =\sqrt{2\delta}\,
    \frac{\Cov(V,h)}{\sqrt{\Var(h)}}+O(\delta),
\end{equation}
which justifies \cref{eq:gamma}.

\paragraph{Bounded polarity preservation.}
Since $g_i\in[0,1]$, $|u_i|\leq1$, and $|\sgn(A_i)|\leq1$, we have $-1\leq g_i\sgn(A_i)u_i\leq1$. Multiplying by $\lambda$ and adding one gives $1-\lambda\leq M_i\leq1+\lambda$. Because $\widehat A_i=A_iM_i$ and $M_i\geq0$, $A_i\widehat A_i=A_i^2M_i\geq0$. Moreover,
\begin{equation}
    |\widehat A_i-A_i|
    =\lambda|A_i|g_i|u_i|
    \leq\lambda|A_i|.
\end{equation}
The normalization in \cref{eq:normalization} is a common nonnegative factor and therefore preserves sign and ordering.

For the normalized distortion, define
\begin{equation}
    \overline M_A=\frac{\sum_j|A_j|M_j}{\sum_j|A_j|},
    \qquad
    \xi_q=\frac{\varepsilon}{\sum_j|A_j|}.
\end{equation}
Whenever $A_i\neq0$, $\widetilde A_i/A_i=M_i/(\overline M_A+\xi_q)$. Both $M_i$ and its $|A|$-weighted average lie in $[1-\lambda,1+\lambda]$, so for $\lambda<1$,
\begin{equation}
    \frac{1-\lambda}{1+\lambda+\xi_q}
    \leq\frac{\widetilde A_i}{A_i}
    \leq\frac{1+\lambda}{1-\lambda+\xi_q}.
    \label{eq:normalized-distortion}
\end{equation}
Within a prompt group, the shared denominator cancels in the ratio of any two realized multipliers $\widetilde A_i/A_i$. Their largest-to-smallest ratio is therefore at most $\kappa(\lambda)=(1+\lambda)/(1-\lambda)$.

\paragraph{Surprisal-ordered response preference.}
For $u>0$, $\partial h_\lambda/\partial u=\lambda|A|(1-\rho_q)>0$; for $u<0$, $\partial h_\lambda/\partial u=\lambda|A|\rho_q>0$. Both branches meet continuously at zero, so $h_\lambda$ is strictly increasing. The normalization in \cref{eq:normalization} multiplies every score in the group by the same positive scalar and preserves the ordering. From \cref{eq:tilt},
\begin{equation}
    \frac{\pi_{\eta,h}(y_i\mid q)}{\pi_{\eta,h}(y_j\mid q)}
    =\frac{\pi_0(y_i\mid q)}{\pi_0(y_j\mid q)}
    e^{\eta(h_i-h_j)},
\end{equation}
which gives \cref{eq:relative-odds} after taking logarithms.

\paragraph{Surprisal alignment and matched-KL derivative.}
Let $\overline B_q=G^{-1}\sum_{i=1}^{G}B_i$ be the group mean of the exploration correction. The empirical covariance expands as
\begin{equation}
\begin{aligned}
\widehat{\Cov}_q(s,B)
&=\frac{1}{G}\sum_i(s_i-\mu_{s,q})(B_i-\overline B_q)\\
&=\frac{1}{G}\sum_i(s_i-\mu_{s,q})B_i,
\end{aligned}
\end{equation}
because centered surprisal sums to zero. Since $u_i$ has the same sign as $s_i-\mu_{s,q}$ and $|A_i|g_i\geq0$, each term $(s_i-\mu_{s,q})B_i=|A_i|g_i(s_i-\mu_{s,q})u_i$ is nonnegative.

For the matched-KL derivative, abbreviate
\begin{equation}
    a=\Cov(V,A),\quad b=\Cov(V,B),\quad
    v=\Var(A),\quad c=\Cov(A,B),\quad d=\Var(B).
\end{equation}
Then
\begin{equation}
    \Gamma_V(A+\lambda B)
    =\frac{a+\lambda b}{\sqrt{v+2\lambda c+\lambda^2d}}.
    \label{eq:gamma-expanded}
\end{equation}
Differentiating at zero gives $(bv-ac)/v^{3/2}$. By \cref{eq:b-perp}, $\Cov(V,B_\perp)=b-ac/v=(bv-ac)/v$, proving \cref{eq:matched-kl-derivative}.

\paragraph{Utilities for correct-mode discovery and diversity.}
Fix a finite partition into correct modes $\mathcal M_q^+$ and write $p_m=\Prob_{Y\sim\pi_0}(Y\in m\mid q)$ and $p_+=\sum_m p_m\in(0,1)$. A mode with mass $p_m(\pi)$ appears at least once in $K$ independent responses with probability $1-(1-p_m(\pi))^K$. Summing these probabilities gives $F_K(\pi)$ in \cref{sec:analysis}. Define the frozen response utility
\[
V_K(y)=
\begin{cases}
K(1-p_{m(y)})^{K-1}, & R(q,y)=1,\\
0, & R(q,y)=0.
\end{cases}
\]
For any differentiable policy path $\pi_\tau$ through $\pi_0$, differentiation gives
\[
\left.\frac{dF_K(\pi_\tau)}{d\tau}\right|_0
=\sum_m K(1-p_m)^{K-1}
  \left.\frac{dp_m(\pi_\tau)}{d\tau}\right|_0
=\left.\frac{d\E_{\pi_\tau}[V_K]}{d\tau}\right|_0.
\]
Consequently, the exponential tilt has discovery slope $\Cov(V_K,h)$ and small-KL gain
\[
F_K(\pi_{\eta_h,h})-F_K(\pi_0)
=\sqrt{2\delta}\,\Gamma_{V_K}(h)+O(\delta).
\]
This is the population quantity estimated by fixed-budget observed correct-mode yield.

For conditional correct-pair diversity, let $w_m=p_m/p_+$, $Q=\sum_m w_m^2$, and $D_{\mathrm{pair}}(\pi_0)=1-Q$. Its frozen response utility is
\[
V_D(y)=
\begin{cases}
\dfrac{2}{p_+}(Q-w_{m(y)}), & R(q,y)=1,\\[4pt]
0, & R(q,y)=0.
\end{cases}
\]
Indeed, writing $\dot p_m=\left.d p_m(\pi_\tau)/d\tau\right|_0$ and $\dot p_+=\sum_m\dot p_m$ gives
\[
\left.\frac{dD_{\mathrm{pair}}(\pi_\tau)}{d\tau}\right|_0
=\frac{2}{p_+}\left(Q\dot p_+-\sum_m w_m\dot p_m\right)
=\left.\frac{d\E_{\pi_\tau}[V_D]}{d\tau}\right|_0.
\]
Its tilt slope is therefore $\Cov(V_D,h)$, and its leading small-KL gain is $\sqrt{2\delta}\,\Gamma_{V_D}(h)$. This identifies the population counterpart of the correct-pair statistic in \cref{app:multi-answer-details}.

With binary rewards and frozen normalization, $A=a_R C+b_R$ for $a_R>0$. Hence $\Cov(C,B_\perp)=0$. Also $\E[V_D]=\E[CV_D]=0$, so $\Cov(V_D,A)=0$ and $\Cov(V_D,B_\perp)=\Cov(V_D,B)$. The discovery utility $V_K$ varies across correct modes for $K\geq2$ whenever their masses differ, and the diversity utility directly measures redistribution within the correct set.

\section{Finite-\texorpdfstring{$\lambda$}{lambda} Matched-KL Certificate}
\label{app:finite-lambda}

The derivative in \cref{eq:matched-kl-derivative} certifies a neighborhood around $\lambda=0$. The following result gives an observable finite interval. Use the abbreviations in \cref{eq:gamma-expanded} and assume $a>0$ and $\Cov(V,B_\perp)>0$. Define
\begin{equation}
    P_0=2a(vb-ac),
    \qquad
    P_1=vb^2-a^2d,
    \label{eq:p0p1}
\end{equation}
\begin{equation}
    \lambda_{\mathrm{num}}=
    \begin{cases}
        +\infty, & b\geq0,\\
        a/(-b), & b<0,
    \end{cases}
    \qquad
    \lambda_{\mathrm{perf}}=
    \begin{cases}
        \min\{\lambda_{\mathrm{num}},P_0/(-P_1)\}, & P_1<0,\\
        \lambda_{\mathrm{num}}, & P_1\geq0.
    \end{cases}
    \label{eq:lambda-perf}
\end{equation}
Then every $0<\lambda<\min\{1,\lambda_{\mathrm{perf}}\}$ satisfying $a+\lambda b>0$ obeys
\begin{equation}
    \Gamma_V(A+\lambda B)>\Gamma_V(A).
    \label{eq:finite-gain}
\end{equation}
To verify the result, both sides of the comparison are positive under $a>0$ and $a+\lambda b>0$. Squaring \cref{eq:finite-gain}, cross-multiplying, cancelling $a^2v$, and dividing by $\lambda>0$ reduces the condition to $P_0+\lambda P_1>0$. The residual-alignment assumption implies $vb-ac=v\Cov(V,B_\perp)>0$, hence $P_0>0$; \cref{eq:lambda-perf} is exactly the interval on which the remaining linear inequality and the positive-numerator condition hold.

For $V=V_K$ in the binary-reward setting above, $a=a_R(1-p_+)\E[V_K]>0$, so the discovery utility satisfies the positive-base-alignment requirement. For conditional diversity, $\Cov(V_D,A)=0$. If $b_D=\Cov(V_D,B)>0$, \cref{eq:gamma-expanded} gives
\[
\Gamma_{V_D}(A+\lambda B)
=\frac{\lambda b_D}{\sqrt{v+2\lambda c+\lambda^2d}}
>\Gamma_{V_D}(A)=0
\]
for every $0<\lambda\leq1$ with $\Var(A+\lambda B)>0$. These conditions compare the leading small-KL coefficients; for each fixed admissible $\lambda$, they yield a positive gain over the base update at sufficiently small KL budgets.

An analogous entropy threshold follows from \cref{eq:entropy-slope}. If $C_B^H=\Cov(S,B)>0$, then the first-order entropy direction is nonnegative for
\begin{equation}
    \lambda\geq\lambda_H
    :=\max\left\{0,-\frac{\Cov(S,A)}{\Cov(S,B)}\right\}.
    \label{eq:lambda-H}
\end{equation}
When the intervals overlap, the measured covariances identify values of $\lambda$ that satisfy both the entropy-direction and matched-KL target-utility conditions.

\section{More Experimental Results}
\label{app:more-experiments}

\subsection{Coefficient Sensitivity and Component Ablations}
\label{app:ablation-results}

\begin{table}[H]
\caption{Phi-3.5-mini coefficient and component ablations at checkpoint 300. Coverage (\%) at $k=16$ for AIME24/25, $k=8$ for Minerva, and $k=4$ otherwise. Full ExPPO, component variants, and the shuffled control use $\lambda=0.2$. Full ExPPO: three-seed mean; other settings: one run. Avg.: unweighted four-task mean; reported macros may differ by 0.01 from rounded task averages. Olympiad/TruthfulQA abbreviate OlympiadBench/TruthfulQA-MC. Bold: highest ExPPO domain mean.}
\label{tab:lambda-ablation}
\centering
\small
\setlength{\tabcolsep}{3.0pt}
\begin{tabular}{@{}clrrrrr@{}}
\toprule
Domain & Setting & AIME24 & AIME25 & Minerva & Olympiad & ID Avg. \\
\midrule
\multirow{13}{*}{ID}
& GRPO & 3.33 & 3.33 & 40.81 & 19.26 & 16.68 \\
\cmidrule(l){2-7}
& $\lambda=0.10$ & 10.00 & 10.00 & 41.54 & 25.19 & 21.68 \\
& $\lambda=0.30$ & 13.33 & 3.33 & 41.91 & 28.74 & 21.83 \\
& $\lambda=0.50$ & 16.67 & 3.33 & 43.38 & 26.81 & 22.55 \\
& $\lambda=0.75$ & 16.67 & 3.33 & 43.38 & 27.85 & 22.81 \\
& $\lambda=1.00$ & 16.67 & 6.67 & 44.49 & 26.96 & 23.70 \\
\cmidrule(l){2-7}
& No final normalization & 10.00 & 10.00 & 40.81 & 27.26 & 22.02 \\
& No pass-rate gate & 16.67 & 3.33 & 43.75 & 27.56 & 22.83 \\
& No $\tanh$ bound & 10.00 & 10.00 & 43.01 & 27.85 & 22.72 \\
& Positive only & 13.33 & 6.67 & 44.49 & 27.85 & 23.09 \\
& Negative only & 10.00 & 6.67 & 42.28 & 26.96 & 21.48 \\
& Within-class shuffled & 10.00 & 6.67 & 42.65 & 28.30 & 21.90 \\
% Appendix audit: retain the full-method reference for direct ablation comparisons;
% its suite means also appear in tab:phi-fgrpo-main, but task entries add detail.
& Full ExPPO & 17.78 & 13.33 & 44.12 & 27.46 & $\mathbf{25.67}$ \\
\midrule
Domain & Setting & GPQA & MMLU-Pro & Safety & TruthfulQA & OOD Avg. \\
\midrule
\multirow{13}{*}{OOD}
& GRPO & 43.08 & 57.14 & 67.50 & 44.88 & 53.15 \\
\cmidrule(l){2-7}
& $\lambda=0.10$ & 47.32 & 67.14 & 71.00 & 64.77 & 62.56 \\
& $\lambda=0.30$ & 54.24 & 72.86 & 77.20 & 76.61 & 70.23 \\
& $\lambda=0.50$ & 52.46 & 67.14 & 77.60 & 77.49 & 68.67 \\
& $\lambda=0.75$ & 57.14 & 75.71 & 77.10 & 76.17 & \textbf{71.53} \\
& $\lambda=1.00$ & 46.65 & 68.57 & 76.80 & 73.25 & 66.32 \\
\cmidrule(l){2-7}
& No final normalization & 50.67 & 67.14 & 75.40 & 70.18 & 65.85 \\
& No pass-rate gate & 46.65 & 65.71 & 76.20 & 75.00 & 65.89 \\
& No $\tanh$ bound & 50.67 & 65.71 & 75.80 & 69.59 & 65.44 \\
& Positive only & 42.86 & 60.00 & 72.40 & 63.74 & 59.75 \\
& Negative only & 46.65 & 72.86 & 74.80 & 69.44 & 65.94 \\
& Within-class shuffled & 46.65 & 61.43 & 74.90 & 70.91 & 63.47 \\
%& Full ExPPO & $48.51{\pm}1.58$ & $69.52{\pm}2.18$ & $75.53{\pm}0.49$ & $71.74{\pm}3.65$ & $66.33{\pm}1.59$ \\

& Full ExPPO & 48.51 & 69.52 & 75.53 & 71.74 & 66.33 \\

\bottomrule
\end{tabular}
\end{table}

\paragraph{All swept coefficients improve coverage over GRPO.}
Every ExPPO coefficient exceeds the matched GRPO baseline on both domain macros. The default $\lambda=0.2$ maximizes ID coverage, while OOD coverage peaks at $\lambda=0.75$, showing a coefficient-dependent adaptation--generalization tradeoff.

\paragraph{Component removal reveals an adaptation--generalization tradeoff.}
Full ExPPO has the highest ID and OOD means among the component variants. Positive-only shaping favors ID coverage (23.09 vs.\ 21.48), whereas negative-only shaping favors OOD coverage (65.94 vs.\ 59.75).

\paragraph{Within-class shuffling tests response-specific credit assignment.}
To test credit assignment beyond class-level reweighting, we compute ExPPO advantages normally, then randomly permute the final sequence-level advantages within each prompt, separately among correct and incorrect responses, before broadcasting them to tokens. This preserves each class's coefficient distribution and total sequence-level mass while disrupting response assignment. Token-weighted update mass can change, so we log it alongside realized policy KL. The shuffled control reaches 21.90 ID, 63.47 OOD, and 42.69 overall coverage, below full ExPPO's three-seed means (25.67, 66.33, and 46.00). This single-run comparison is consistent with a benefit from response-specific assignment; token-weighted mass and realized KL provide additional factors for interpreting the result.

\subsection{Closest Baselines at the Phi-3.5-mini Ablation Budget}
\label{app:phi-fgrpo-ablation}

\begin{table}[H]
\caption{Phi-3.5-mini at checkpoint 300 under the ablation setting of \cref{tab:lambda-ablation}. Entries are mean $\pm$ sample standard deviation across three training seeds per method. Avg.: unweighted four-task mean within each seed; Overall: eight-task mean. Task summaries use the reported seed scores; macro summaries retain author-reported precision, which can differ by 0.01 point from aggregation of rounded task entries. Individual seeds: \cref{tab:phi-fgrpo-seeds,tab:phi-ur-repo-seeds}. Bold: highest mean across the four methods. ExPPO uses $\lambda=0.2$.}
\label{tab:phi-fgrpo-ablation}
\centering
\small
\setlength{\tabcolsep}{3.0pt}
\begin{tabular*}{\linewidth}{@{\extracolsep{\fill}}clrrrrr@{}}
\toprule
\multicolumn{7}{@{}l}{\textbf{A. Coverage (\%)}} \\
\midrule
Domain & Benchmark & $k$ & F-GRPO & GRPO-UR & REPO-R & ExPPO \\
\midrule
\multirow{5}{*}{ID} & AIME24 & 16 & $5.55{\pm}6.94$ & $12.22{\pm}1.92$ & $14.44{\pm}5.09$ & $\mathbf{17.78{\pm}3.85}$ \\
 & AIME25 & 16 & $3.33{\pm}3.34$ & $7.78{\pm}5.09$ & $2.22{\pm}1.92$ & $\mathbf{13.33{\pm}0.00}$ \\
 & Minerva & 8 & $33.58{\pm}10.99$ & $41.79{\pm}2.50$ & $\mathbf{46.20{\pm}1.12}$ & $44.12{\pm}0.97$ \\
 & Olympiad & 4 & $13.04{\pm}3.71$ & $24.84{\pm}1.77$ & $24.89{\pm}0.40$ & $\mathbf{27.46{\pm}0.68}$ \\
\cmidrule(l){2-7}
 & Avg. & --- & $13.88{\pm}5.76$ & $21.66{\pm}0.75$ & $21.94{\pm}1.89$ & $\mathbf{25.67{\pm}1.31}$ \\
\midrule
\multirow{5}{*}{OOD} & GPQA & 4 & $33.78{\pm}18.91$ & $45.54{\pm}4.84$ & $\mathbf{50.60{\pm}0.25}$ & $48.51{\pm}1.58$ \\
 & MMLU-Pro & 4 & $41.91{\pm}27.86$ & $68.10{\pm}6.44$ & $67.62{\pm}1.65$ & $\mathbf{69.52{\pm}2.18}$ \\
 & Safety & 4 & $55.73{\pm}24.30$ & $70.53{\pm}5.23$ & $75.30{\pm}0.26$ & $\mathbf{75.53{\pm}0.49}$ \\
 & TruthfulQA & 4 & $40.20{\pm}30.83$ & $66.86{\pm}5.77$ & $\mathbf{71.88{\pm}0.75}$ & $71.74{\pm}3.65$ \\
\cmidrule(l){2-7}
 & Avg. & --- & $42.91{\pm}25.37$ & $62.76{\pm}5.48$ & $\mathbf{66.35{\pm}0.38}$ & $66.33{\pm}1.59$ \\
\midrule
Overall & Avg. & --- & $28.39{\pm}15.22$ & $42.21{\pm}2.65$ & $44.14{\pm}1.00$ & $\mathbf{46.00{\pm}1.39}$ \\
\midrule
\multicolumn{7}{@{}l}{\textbf{B. Response accuracy (\%)}} \\
\midrule
Domain & Benchmark & $k$ & F-GRPO & GRPO-UR & REPO-R & ExPPO \\
\midrule
\multirow{5}{*}{ID} & AIME24 & 16 & $1.74{\pm}1.67$ & $2.02{\pm}0.64$ & $1.53{\pm}0.44$ & $\mathbf{2.85{\pm}0.43}$ \\
 & AIME25 & 16 & $0.21{\pm}0.21$ & $0.76{\pm}0.12$ & $0.14{\pm}0.12$ & $\mathbf{1.11{\pm}0.32}$ \\
 & Minerva & 8 & $17.55{\pm}3.72$ & $22.96{\pm}2.71$ & $\mathbf{24.25{\pm}0.57}$ & $23.05{\pm}0.18$ \\
 & Olympiad & 4 & $6.18{\pm}0.61$ & $13.42{\pm}2.33$ & $12.65{\pm}0.47$ & $\mathbf{15.36{\pm}0.54}$ \\
\cmidrule(l){2-7}
 & Avg. & --- & $6.42{\pm}0.81$ & $9.79{\pm}1.11$ & $9.64{\pm}0.19$ & $\mathbf{10.59{\pm}0.27}$ \\
\midrule
\multirow{5}{*}{OOD} & GPQA & 4 & $21.95{\pm}10.03$ & $\mathbf{27.71{\pm}1.27}$ & $26.69{\pm}0.17$ & $25.17{\pm}0.79$ \\
 & MMLU-Pro & 4 & $34.76{\pm}23.49$ & $\mathbf{52.50{\pm}2.86}$ & $46.91{\pm}0.83$ & $48.57{\pm}2.70$ \\
 & Safety & 4 & $51.58{\pm}25.65$ & $68.12{\pm}5.62$ & $\mathbf{72.10{\pm}0.16}$ & $71.98{\pm}0.20$ \\
 & TruthfulQA & 4 & $38.00{\pm}30.22$ & $63.09{\pm}5.76$ & $62.96{\pm}1.16$ & $\mathbf{63.32{\pm}1.85}$ \\
\cmidrule(l){2-7}
 & Avg. & --- & $36.57{\pm}22.19$ & $\mathbf{52.85{\pm}3.62}$ & $52.16{\pm}0.15$ & $52.26{\pm}1.17$ \\
\midrule
Overall & Avg. & --- & $21.50{\pm}11.43$ & $31.32{\pm}2.37$ & $30.90{\pm}0.08$ & $\mathbf{31.43{\pm}0.72}$ \\
\bottomrule
\end{tabular*}
\end{table}

\paragraph{Coverage and accuracy favor different allocations.}
ExPPO has the highest overall coverage (46.00), followed by REPO-R (44.14), GRPO-UR (42.21), and F-GRPO (28.39). ExPPO leads coverage on five benchmarks, while REPO-R leads on Minerva, GPQA, and TruthfulQA. GRPO-UR combines the highest OOD response accuracy with lower OOD coverage, illustrating the distinct roles of single-response success and prompt coverage at the reported sampling budgets. REPO-R has the smallest OOD variation in both metrics; F-GRPO has the largest. These are descriptive comparisons across three training seeds.

\begin{table}[!htbp]
\caption{Individual F-GRPO and ExPPO training seeds for Phi-3.5-mini at checkpoint 300. Each cell is coverage / response accuracy (percent), at $k=16$ for AIME24/25, 8 for Minerva, and 4 otherwise. Task scores and macro rows retain author-reported precision; aggregation from rounded task entries can differ by 0.01 point. The other six runs appear in \cref{tab:phi-ur-repo-seeds}.}
\label{tab:phi-fgrpo-seeds}
\centering
\small
\setlength{\tabcolsep}{4pt}
\begin{tabular*}{\linewidth}{@{\extracolsep{\fill}}llrrr@{}}
\toprule
Method & Benchmark / suite & Seed 42 & Seed 43 & Seed 44 \\
\midrule
\multirow{11}{*}{F-GRPO} & AIME24 & 3.33 / 3.33 & 0.00 / 0.00 & 13.33 / 1.88 \\
 & AIME25 & 0.00 / 0.00 & 3.33 / 0.21 & 6.67 / 0.42 \\
 & OlympiadBench & 9.19 / 6.19 & 13.33 / 5.56 & 16.59 / 6.78 \\
 & Minerva & 21.69 / 13.28 & 35.66 / 19.30 & 43.38 / 20.08 \\
 & \textbf{ID macro} & \textbf{8.55 / 5.70} & \textbf{13.08 / 6.27} & \textbf{19.99 / 7.29} \\
\cmidrule(l){2-5}
 & MMLU-Pro & 10.00 / 8.21 & 54.29 / 43.21 & 61.43 / 52.86 \\
 & Safety & 27.70 / 22.10 & 70.80 / 68.75 & 68.70 / 63.90 \\
 & TruthfulQA-MC & 4.82 / 3.36 & 54.53 / 51.68 & 61.26 / 58.95 \\
 & GPQA & 12.50 / 10.38 & 40.18 / 27.29 & 48.66 / 28.18 \\
 & \textbf{OOD macro} & \textbf{13.76 / 11.01} & \textbf{54.95 / 47.73} & \textbf{60.01 / 50.97} \\
\cmidrule(l){2-5}
 & \textbf{Overall macro} & \textbf{11.15 / 8.36} & \textbf{34.02 / 27.00} & \textbf{40.00 / 29.13} \\
\midrule
\multirow{11}{*}{ExPPO} & AIME24 & 20.00 / 3.33 & 13.33 / 2.71 & 20.00 / 2.50 \\
 & AIME25 & 13.33 / 1.46 & 13.33 / 0.83 & 13.33 / 1.04 \\
 & OlympiadBench & 27.85 / 15.96 & 26.67 / 15.19 & 27.85 / 14.93 \\
 & Minerva & 45.22 / 22.89 & 43.38 / 23.02 & 43.75 / 23.25 \\
 & \textbf{ID macro} & \textbf{26.60 / 10.91} & \textbf{24.18 / 10.44} & \textbf{26.23 / 10.43} \\
\cmidrule(l){2-5}
 & MMLU-Pro & 70.00 / 51.43 & 67.14 / 46.07 & 71.43 / 48.21 \\
 & Safety & 76.10 / 71.93 & 75.30 / 72.20 & 75.20 / 71.80 \\
 & TruthfulQA-MC & 75.88 / 65.17 & 69.01 / 63.30 & 70.32 / 61.48 \\
 & GPQA & 50.22 / 25.89 & 48.21 / 25.28 & 47.10 / 24.33 \\
 & \textbf{OOD macro} & \textbf{68.05 / 53.60} & \textbf{64.92 / 51.71} & \textbf{66.01 / 51.46} \\
\cmidrule(l){2-5}
 & \textbf{Overall macro} & \textbf{47.33 / 32.26} & \textbf{44.55 / 31.08} & \textbf{46.12 / 30.94} \\
\bottomrule
\end{tabular*}
\end{table}

\FloatBarrier
\begin{table}[H]
\caption{Individual GRPO-UR and REPO-R training seeds for Phi-3.5-mini at checkpoint 300. Each cell is coverage / response accuracy (percent), with budgets and rounding conventions as in \cref{tab:phi-fgrpo-seeds}. Together, the two seed tables report all twelve runs.}
\label{tab:phi-ur-repo-seeds}
\centering
\small
\setlength{\tabcolsep}{4pt}
\begin{tabular*}{\linewidth}{@{\extracolsep{\fill}}llrrr@{}}
\toprule
Method & Benchmark / suite & Seed 42 & Seed 43 & Seed 44 \\
\midrule
\multirow{11}{*}{GRPO-UR} & AIME24 & 13.33 / 1.88 & 10.00 / 1.46 & 13.33 / 2.71 \\
 & AIME25 & 6.67 / 0.83 & 3.33 / 0.63 & 13.33 / 0.83 \\
 & OlympiadBench & 25.63 / 15.00 & 26.07 / 14.52 & 22.81 / 10.74 \\
 & Minerva & 42.65 / 25.18 & 43.75 / 23.76 & 38.97 / 19.94 \\
 & \textbf{ID macro} & \textbf{22.07 / 10.72} & \textbf{20.79 / 10.09} & \textbf{22.11 / 8.56} \\
\cmidrule(l){2-5}
 & MMLU-Pro & 74.29 / 55.36 & 68.57 / 52.50 & 61.43 / 49.64 \\
 & Safety & 73.80 / 71.55 & 73.30 / 71.18 & 64.50 / 61.63 \\
 & TruthfulQA-MC & 72.08 / 67.65 & 67.84 / 64.99 & 60.67 / 56.62 \\
 & GPQA & 48.66 / 29.07 & 47.99 / 26.56 & 39.96 / 27.51 \\
 & \textbf{OOD macro} & \textbf{67.21 / 55.91} & \textbf{64.42 / 53.81} & \textbf{56.64 / 48.85} \\
\cmidrule(l){2-5}
 & \textbf{Overall macro} & \textbf{44.64 / 33.32} & \textbf{42.61 / 31.95} & \textbf{39.38 / 28.70} \\
\midrule
\multirow{11}{*}{REPO-R} & AIME24 & 13.33 / 1.88 & 10.00 / 1.04 & 20.00 / 1.67 \\
 & AIME25 & 3.33 / 0.21 & 0.00 / 0.00 & 3.33 / 0.21 \\
 & OlympiadBench & 24.44 / 12.22 & 25.04 / 12.59 & 25.19 / 13.15 \\
 & Minerva & 45.22 / 23.62 & 45.96 / 24.72 & 47.43 / 24.40 \\
 & \textbf{ID macro} & \textbf{21.58 / 9.48} & \textbf{20.25 / 9.59} & \textbf{23.99 / 9.86} \\
\cmidrule(l){2-5}
 & MMLU-Pro & 68.57 / 46.43 & 65.71 / 46.43 & 68.57 / 47.86 \\
 & Safety & 75.50 / 72.28 & 75.40 / 71.98 & 75.00 / 72.03 \\
 & TruthfulQA-MC & 72.51 / 63.89 & 72.08 / 63.34 & 71.05 / 61.66 \\
 & GPQA & 50.45 / 26.73 & 50.89 / 26.51 & 50.45 / 26.84 \\
 & \textbf{OOD macro} & \textbf{66.76 / 52.33} & \textbf{66.02 / 52.06} & \textbf{66.27 / 52.10} \\
\cmidrule(l){2-5}
 & \textbf{Overall macro} & \textbf{44.17 / 30.91} & \textbf{43.13 / 30.83} & \textbf{45.13 / 30.98} \\
\bottomrule
\end{tabular*}
\end{table}

\paragraph{Seed ranges clarify the robustness of the comparison.}
Every ExPPO ID coverage run (24.18--26.60) exceeds every baseline run; the best baseline run is REPO-R at 23.99. OOD coverage ranges overlap for ExPPO (64.92--68.05), REPO-R (66.02--66.76), and GRPO-UR (56.64--67.21). F-GRPO spans 13.76--60.01, with its lowest run accounting for much of its variation. GRPO-UR's higher mean OOD response accuracy comes with a wider range (48.85--55.91) than ExPPO (51.46--53.60) or REPO-R (52.06--52.33).

\FloatBarrier
\subsection{Independent-Run Stability on Phi-3.5-mini}
\label{app:phi-independent-runs}

\begin{table}[H]
\caption{Phi-3.5-mini performance across the original and two additional independently trained runs at the last checkpoint. Domain: In-Domain (ID) or Out-of-Domain (OOD); upper panel: any-success coverage; lower panel: mean verified-response share. Scores are in percentage points at $k=16$ for AIME24/25, $k=8$ for Minerva, and $k=4$ otherwise. Avg.: unweighted four-task mean within each run, followed by the across-run mean $\pm$ sample standard deviation. Olympiad/TruthfulQA abbreviate OlympiadBench/TruthfulQA-MC. Bold: stronger mean per column.}
\label{tab:phi-independent-runs}
\centering
\scriptsize
\setlength{\tabcolsep}{2.8pt}
\begin{tabular}{@{}clrrrrr@{}}
\toprule
\multicolumn{7}{c}{Coverage} \\
\midrule
Domain & Method & AIME24 & AIME25 & Minerva & Olympiad & Avg. \\
\midrule
\multirow{2}{*}{ID}
& GRPO & $7.78{\pm}5.09$ & $2.22{\pm}1.92$ & $33.09{\pm}13.78$ & $16.15{\pm}7.40$ & $14.81{\pm}6.89$ \\
& ExPPO & $\mathbf{14.44{\pm}5.09}$ & $\mathbf{8.89{\pm}3.85}$ & $\mathbf{44.85{\pm}2.24}$ & $\mathbf{27.80{\pm}1.11}$ & $\mathbf{24.00{\pm}0.62}$ \\
\midrule
Domain & Method & GPQA & MMLU-Pro & Safety & TruthfulQA & Avg. \\
\midrule
\multirow{2}{*}{OOD}
& GRPO & $41.00{\pm}6.19$ & $58.57{\pm}6.23$ & $70.53{\pm}4.37$ & $64.57{\pm}2.05$ & $58.67{\pm}3.07$ \\
& ExPPO & $\mathbf{49.78{\pm}3.72}$ & $\mathbf{71.90{\pm}5.02}$ & $\mathbf{75.63{\pm}0.67}$ & $\mathbf{71.93{\pm}3.98}$ & $\mathbf{67.31{\pm}3.25}$ \\
\bottomrule
\end{tabular}

\vspace{5pt}
\begin{tabular}{@{}clrrrrr@{}}
\toprule
\multicolumn{7}{c}{Response accuracy} \\
\midrule
Domain & Method & AIME24 & AIME25 & Minerva & Olympiad & Avg. \\
\midrule
\multirow{2}{*}{ID}
& GRPO & $1.60{\pm}1.53$ & $0.21{\pm}0.21$ & $18.03{\pm}6.49$ & $8.04{\pm}3.06$ & $6.97{\pm}2.13$ \\
& ExPPO & $\mathbf{2.71{\pm}0.55}$ & $\mathbf{0.76{\pm}0.12}$ & $\mathbf{24.32{\pm}2.06}$ & $\mathbf{15.42{\pm}0.64}$ & $\mathbf{10.80{\pm}0.82}$ \\
\midrule
Domain & Method & GPQA & MMLU-Pro & Safety & TruthfulQA & Avg. \\
\midrule
\multirow{2}{*}{OOD}
& GRPO & $\mathbf{26.94{\pm}1.18}$ & $\mathbf{49.40{\pm}2.58}$ & $67.16{\pm}5.56$ & $62.17{\pm}2.08$ & $51.42{\pm}0.73$ \\
& ExPPO & $25.39{\pm}1.12$ & $47.62{\pm}1.35$ & $\mathbf{72.12{\pm}0.28}$ & $\mathbf{63.46{\pm}2.07}$ & $\mathbf{52.15{\pm}0.98}$ \\
\bottomrule
\end{tabular}
\end{table}

\paragraph{ExPPO is stable across mathematical adaptation and OOD generalization.}
ExPPO raises mean coverage on all eight tasks and both domain averages. Its ID coverage varies little across training runs, while GRPO is more sensitive to the training trajectory. Response accuracy follows the same macro pattern: ExPPO gains on the mathematical suite and the OOD average, with complementary task-level strengths on the OOD suite.

\subsection{Code-Reasoning Adaptation and OOD Generalization}
\label{app:code-reasoning}

We evaluate Phi-3.5-mini on a CodeReasoner-derived Python execution-reasoning set using the released data\footnote{\href{https://huggingface.co/datasets/just-for-paper-submission/CodeReasoner-Dataset}{CodeReasoner-Dataset on Hugging Face}} \citep{tang2025codereasoner}. \Cref{tab:code-main} reports the code and OOD-summary results; \cref{tab:code-reasoning} gives the OOD task breakdown for the same evaluations.

\begin{table}[H]
\caption{OOD generalization after code adaptation, disaggregating \cref{tab:code-main}. Upper panel: any-success prompt coverage at $k=4$; lower panel: mean verified-response share. Scores are percentages. Initialization: fixed base model; trained entries: mean $\pm$ sample standard deviation across three independent runs at the last checkpoint. Bold: best trained mean.}
\label{tab:code-reasoning}
\centering
\small
\setlength{\tabcolsep}{4pt}
\begin{tabular*}{\linewidth}{@{\extracolsep{\fill}}lrrrr@{}}
\toprule
\multicolumn{5}{c}{Coverage} \\
\midrule
Method & GPQA & MMLU-Pro & Safety & TruthfulQA-MC \\
\midrule
Initialization & 38.17 & 58.57 & 74.50 & 66.37 \\
GRPO & $32.96{\pm}0.46$ & $47.62{\pm}1.65$ & $70.10{\pm}0.30$ & $58.28{\pm}0.83$ \\
GRPO-H & $37.57{\pm}0.90$ & $59.52{\pm}3.60$ & $72.37{\pm}0.40$ & $72.56{\pm}0.97$ \\
ExPPO & $\mathbf{44.35{\pm}1.65}$ & $\mathbf{65.24{\pm}4.12}$ & $\mathbf{76.37{\pm}0.25}$ & $\mathbf{77.00{\pm}0.69}$ \\
\midrule
\multicolumn{5}{c}{Response accuracy} \\
\midrule
Initialization & 17.52 & 33.93 & 66.75 & 49.52 \\
GRPO & $14.55{\pm}0.23$ & $30.60{\pm}3.38$ & $61.88{\pm}0.65$ & $38.65{\pm}0.93$ \\
GRPO-H & $16.39{\pm}0.67$ & $35.71{\pm}1.07$ & $60.65{\pm}0.90$ & $55.67{\pm}0.31$ \\
ExPPO & $\mathbf{21.73{\pm}0.90}$ & $\mathbf{44.64{\pm}4.02}$ & $\mathbf{72.51{\pm}0.14}$ & $\mathbf{65.22{\pm}0.77}$ \\
\bottomrule
\end{tabular*}
\end{table}

\paragraph{ExPPO improves code adaptation and OOD generalization.}
ExPPO finishes highest on both code metrics and every OOD column. Its higher OOD response accuracy shows that the generalization gain includes a larger share of correct responses across samples.

\subsection{Per-Benchmark Checkpoint Trajectories}
\label{app:per-benchmark-trends}

\begin{figure}[H]
    \centering
    \includegraphics[width=\textwidth]{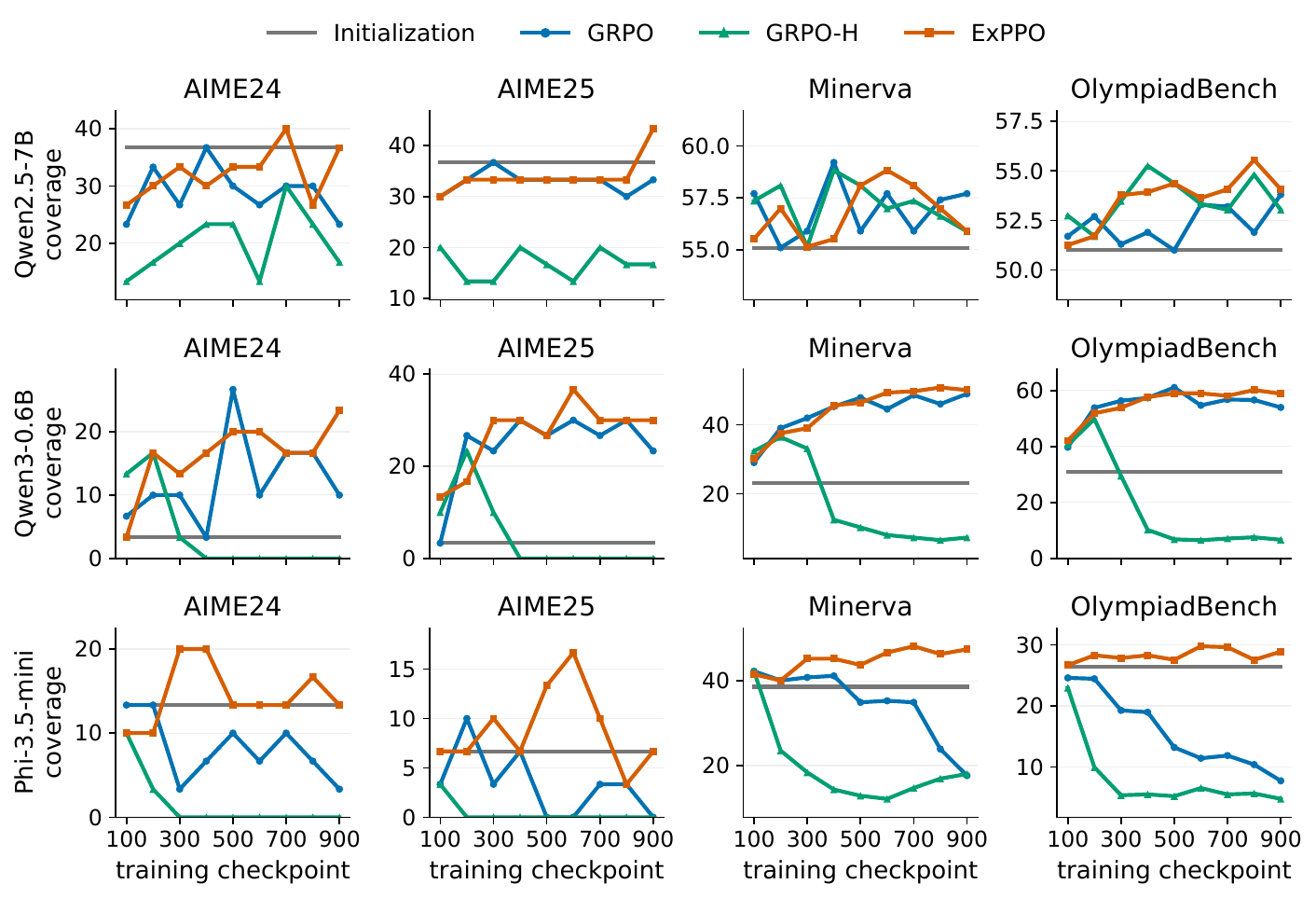}
    \caption{Per-task In-Domain (ID) coverage through the last checkpoint; disaggregation of \cref{fig:training-macro}. Rows: Qwen2.5-7B, Qwen3-0.6B, Phi-3.5-mini; columns: AIME24, AIME25, Minerva, OlympiadBench. Horizontal axis: checkpoint; vertical axis: coverage in percentage points at $k=16,16,8,4$, respectively. Gray lines: Initialization; colored curves: GRPO, GRPO-H, ExPPO.}
    \label{fig:training-id}
\end{figure}

\begin{figure}[H]
    \centering
    \includegraphics[width=\textwidth]{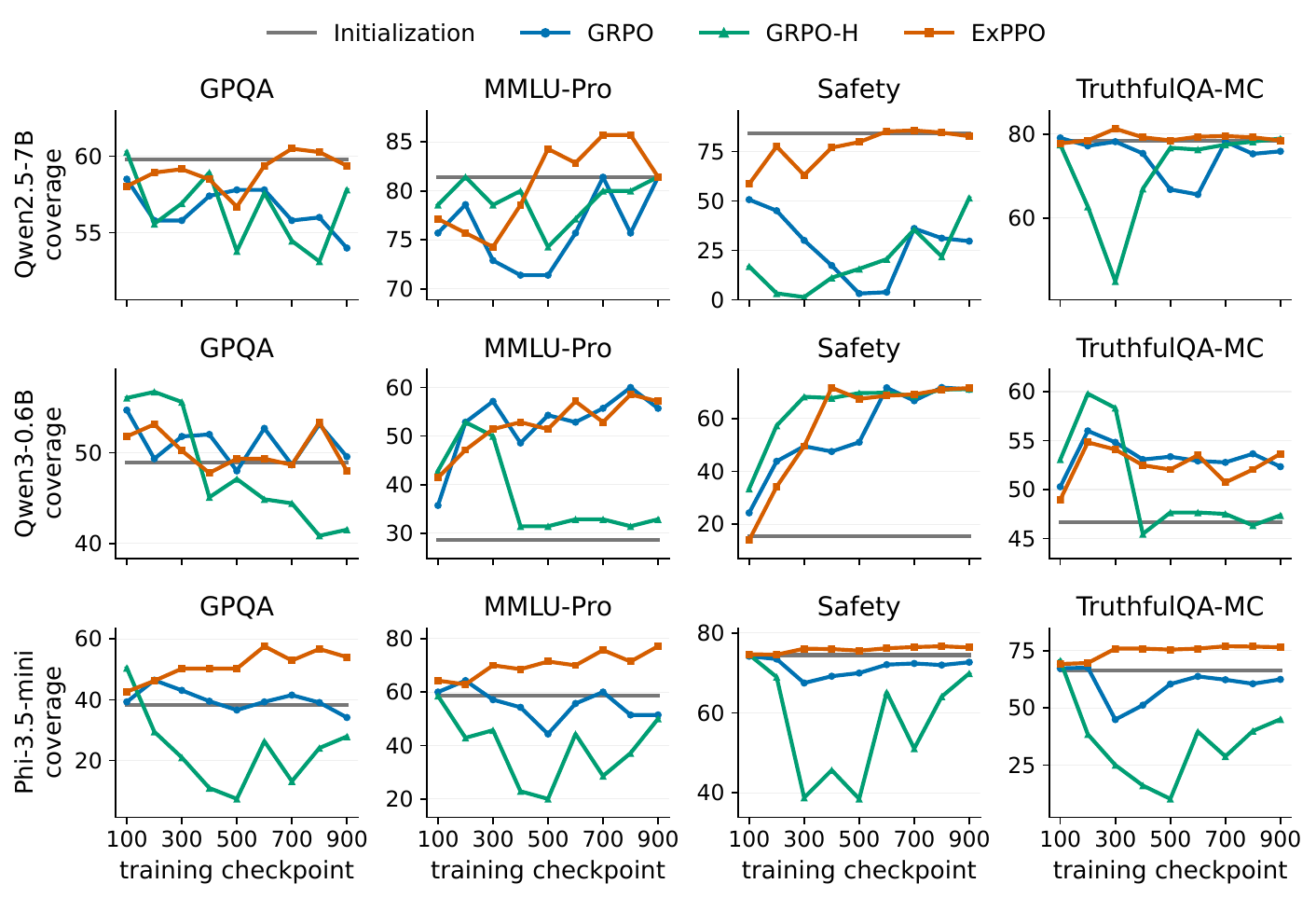}
    \caption{Per-task Out-of-Domain (OOD) coverage through the last checkpoint; disaggregation of \cref{fig:training-macro}. Rows: Qwen2.5-7B, Qwen3-0.6B, Phi-3.5-mini; columns: GPQA, MMLU-Pro, Safety, TruthfulQA-MC. Horizontal axis: checkpoint; vertical axis: coverage in percentage points at $k=4$. Gray lines: Initialization; colored curves: GRPO, GRPO-H, ExPPO.}
    \label{fig:training-ood}
\end{figure}

\paragraph{ExPPO is consistent while GRPO-H collapses on smaller backbones.}
\Cref{fig:training-id,fig:training-ood} recover the macro pattern while exposing task-specific variation. ExPPO finishes strongest on most panels, whereas GRPO-H loses mathematical coverage across Qwen3-0.6B and Phi-3.5-mini and becomes unstable on several OOD tasks.

\FloatBarrier

\subsection{DAPO-Based ExPPO}
\label{app:dapo-results}

\begin{figure}[H]
    \centering
    \includegraphics[width=0.95\linewidth]{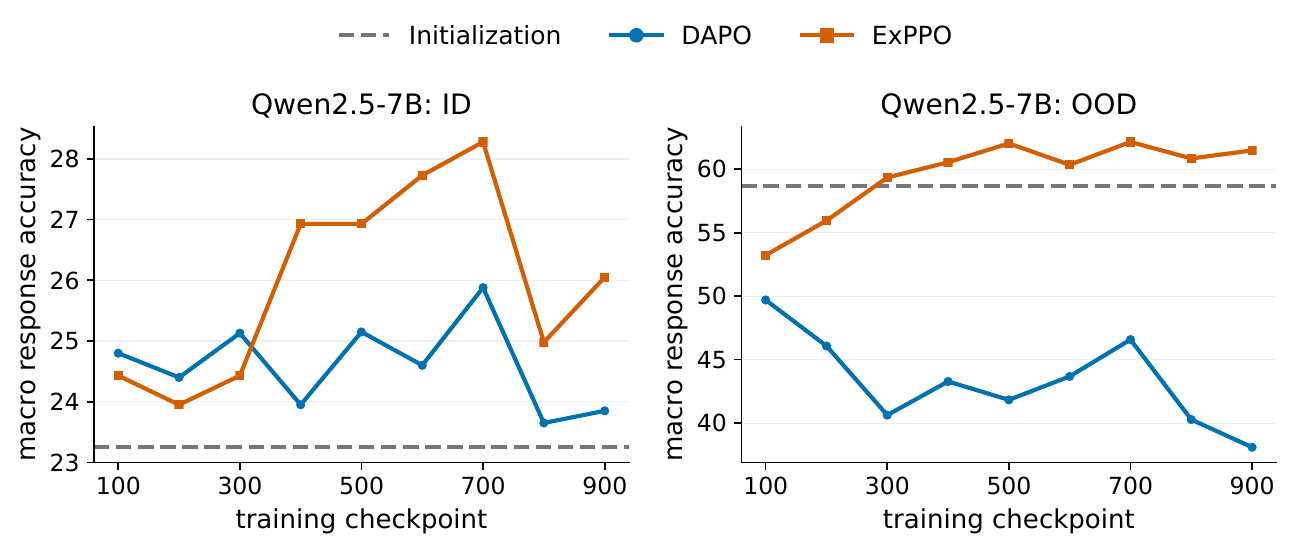}
    \caption{Qwen2.5-7B macro response-accuracy trajectories under DAPO. Left: unweighted mean over AIME24, AIME25, Minerva, OlympiadBench (In-Domain; ID); right: unweighted mean over GPQA, MMLU-Pro, Safety, TruthfulQA-MC (Out-of-Domain; OOD). Horizontal axis: checkpoint; vertical axis: response accuracy in percentage points. Gray dashed lines: Initialization; colored curves: DAPO, ExPPO.}
    \label{fig:dapo-accuracy}
\end{figure}

\paragraph{ExPPO extends to the DAPO recipe.}
We apply ExPPO to DAPO while retaining asymmetric clipping, dynamic sampling, token-level loss aggregation, and overlong-response shaping \citep{yu2025dapo}. \Cref{fig:dapo-accuracy} shows DAPO's OOD macro declining through training, whereas ExPPO rises and finishes above its initialization.

\begin{table}[H]
\caption{Qwen2.5-7B DAPO response accuracy at the last checkpoint. Upper/lower panels: In-Domain (ID)/Out-of-Domain (OOD). Values: mean verified-response share in percentage points at $k=16$ for AIME24/25, $k=8$ for Minerva, and $k=4$ otherwise. Avg.: unweighted four-task mean. Olympiad/TruthfulQA abbreviate OlympiadBench/TruthfulQA-MC. Initialization: fixed base model; bold: better trained method per column.}
\label{tab:dapo-accuracy}
\centering
\small
\setlength{\tabcolsep}{4.5pt}
\begin{tabular}{lrrrrr}
\toprule
\multicolumn{6}{c}{ID} \\
\cmidrule(lr){1-6}
Method & AIME24 & AIME25 & Minerva & Olympiad & Avg. \\
\midrule
Initialization & 12.10 & 7.70 & 37.10 & 36.10 & 23.25 \\
\midrule
DAPO & 10.42 & 7.92 & \textbf{41.31} & 35.78 & 23.85 \\
ExPPO & \textbf{14.38} & \textbf{10.21} & 40.85 & \textbf{38.67} & \textbf{26.05} \\
\bottomrule
\end{tabular}

\vspace{4pt}
\begin{tabular}{lrrrrr}
\toprule
\multicolumn{6}{c}{OOD} \\
\cmidrule(lr){1-6}
Method & GPQA & MMLU-Pro & Safety & TruthfulQA & Avg. \\
\midrule
Initialization & 32.80 & 62.50 & 73.00 & 66.40 & 58.68 \\
\midrule
DAPO & 33.04 & 64.64 & 0.85 & 53.91 & 38.10 \\
ExPPO & \textbf{36.10} & \textbf{71.79} & \textbf{72.03} & \textbf{66.01} & \textbf{61.48} \\
\bottomrule
\end{tabular}
\end{table}

\paragraph{ExPPO restores OOD generalization under DAPO.}
In \cref{tab:dapo-accuracy}, ExPPO raises both split macros over DAPO, exceeds the initialization on both, and leads DAPO on seven of eight tasks. The largest recovery occurs on Safety, where DAPO loses most of its initial accuracy and ExPPO preserves it.

\subsection{Controlled Multi-Answer Task}
\label{app:multi-answer-details}

\paragraph{Exact verification makes correct-support breadth measurable.}
The controlled set is a deterministically generated synthetic evaluation panel with 64 fixed prompts, 4--46 exhaustively verified modes per prompt in the realized bank, and zero exact prompt overlap with DAPO-Math-17k. Each prompt specifies four positive denominations $d=(d_1,d_2,d_3,d_4)$ sampled from 2--13, a target $T$ sampled from 24--70, and a maximum count of six. The model is instructed to return a JSON object of the form \texttt{\{"counts":[c1,c2,c3,c4]\}}. The unified final-answer parser selects the final-most literal JSON \texttt{"counts"} field in the response. A response is correct exactly when this field contains four integers satisfying
\begin{equation}
    c_j\in\{0,1,\ldots,6\}\quad\text{for every }j,
    \qquad \sum_{j=1}^{4}c_jd_j=T.
    \label{eq:coin-change-rule}
\end{equation}
Each distinct verified count vector is a semantic answer mode. For example, when $d=(4,9,10,13)$ and $T=41$, four valid modes are
\begin{equation}
    (0,2,1,1),\quad (1,3,1,0),\quad (2,0,2,1),\quad (3,1,2,0).
    \label{eq:coin-change-example}
\end{equation}
This construction gives exact correctness and mode identity.

Every model--method pair receives the same prompt bank and $K=128$ samples per prompt using temperature 1.0, top-$p=1.0$, and top-$k=0$. Each prompt contains the user task and JSON-output instruction and is rendered with the tokenizer's native template; Qwen3 thinking is disabled. Evaluation caps are 2,048 tokens for Qwen2.5-7B and Phi-3.5-mini and 8,192 tokens for Qwen3-0.6B. Qwen2.5-7B and Qwen3-0.6B are generated directly at these caps; the Phi-3.5-mini result is exact 2,048-token prefix re-verification of a common 16,384-token response bank.

For prompt $q$, let $n_{q,m}$ be the number of verified responses in correct mode $m$, $C_q=\sum_m n_{q,m}$, and $U_q=|\{m:n_{q,m}>0\}|$. Fixed-budget observed correct-mode yield is $64^{-1}\sum_q U_q$. When $C_q>0$, let $p_{q,m}=n_{q,m}/C_q$ and define
\begin{equation}
    N_{\mathrm{eff}}(q)=
    \exp\!\left(-\sum_m p_{q,m}\log p_{q,m}\right).
    \label{eq:effective-modes}
\end{equation}
We set $N_{\mathrm{eff}}(q)=0$ for unsolved prompts, so the prompt mean is a fixed-budget effective correct-mode yield that combines discoverability and diversity. \Cref{tab:multi-answer-fixed-full} supplements the main results with response counts and diversity-eligible prompt counts.

\begin{table}[H]
\caption{Additional controlled-task diagnostics ($K=128$; 64 prompts; last checkpoint). Correct: verified responses among 8,192 samples; $C\!\geq\!2$: prompts with at least two correct responses, eligible for diversity evaluation. Qwen2.5-7B ExPPO uses the earlier evaluation underlying the GRPO/GRPO-H pairs in \cref{tab:multi-answer} and the GRPO-UR comparison in \cref{tab:multi-answer-nearest}. Other rows match the main evaluation.}
\label{tab:multi-answer-fixed-full}
\centering
\small
\setlength{\tabcolsep}{5pt}
\begin{tabular}{llrr}
\toprule
Model & Method & Correct / 8,192 & $C\!\geq\!2$ / 64 \\
\midrule
\multirow{3}{*}{Qwen2.5-7B}
& GRPO & 528 & 51 \\
& GRPO-H & 793 & 55 \\
& ExPPO & 2,608 & 64 \\
\midrule
\multirow{3}{*}{Qwen3-0.6B}
& GRPO & 8,063 & 64 \\
& GRPO-H & 2 & 1 \\
& ExPPO & 8,007 & 64 \\
\midrule
\multirow{3}{*}{Phi-3.5-mini}
& GRPO & 74 & 13 \\
& GRPO-H & 41 & 7 \\
& ExPPO & 185 & 24 \\
\bottomrule
\end{tabular}
\end{table}

To isolate the distribution among correct responses, for prompts with $C_q\geq2$ we compute correct-pair diversity
\begin{equation}
    D_q
    =1-\frac{\sum_m n_{q,m}(n_{q,m}-1)}{C_q(C_q-1)}.
    \label{eq:correct-pair-diversity}
\end{equation}
This is the probability that two verified-correct responses sampled without replacement from the observed bank have different semantic modes. For each ExPPO--baseline comparison, we average $D_q$ equally over the common prompts where both methods have at least two correct responses. The intervals use 10,000 paired prompt-bootstrap resamples. The main table reports the Qwen2.5-7B GRPO, GRPO-H, F-GRPO, and REPO-R pairs and the Qwen3-0.6B GRPO pair; \cref{tab:multi-answer-pair-full} gives the remaining core comparisons, and \cref{tab:multi-answer-nearest} adds GRPO-UR.

\begin{table}[H]
\caption{Additional core-baseline correct-pair comparisons from the evaluations in \cref{tab:multi-answer-fixed-full}; other pairs appear in \cref{tab:multi-answer}. Shared: prompts with at least two correct responses per method. $D$: mean distinct-mode probability of correct pairs on that shared set; $\Delta D$: ExPPO minus baseline; CI: paired prompt-bootstrap 95\% interval. $D$ uses a 0--1 scale, with differences and intervals in the same units, versus the main table's 0--100 scale. The one-prompt Qwen3-0.6B comparison is reported as a point estimate.}
\label{tab:multi-answer-pair-full}
\centering
\scriptsize
\setlength{\tabcolsep}{4.0pt}
\begin{tabular}{llrrrrr}
\toprule
Model & Baseline & Shared / 64 & Baseline $D$ & ExPPO $D$ & $\Delta D$ & CI \\
\midrule
% Appendix audit 2026-09-08: these pairs duplicate tab:multi-answer after scaling by 100.
% Qwen2.5-7B & GRPO   & 51 & 0.3398 & 0.5057 & \textbf{$+0.1659$} & \textbf{$[+0.0619,+0.2710]$} \\
%             & GRPO-H & 55 & 0.3514 & 0.4877 & \textbf{$+0.1363$} & \textbf{$[+0.0537,+0.2171]$} \\
% \midrule
% Qwen3-0.6B & GRPO   & 64 & 0.5146 & 0.5844 & \textbf{$+0.0698$} & \textbf{$[+0.0301,+0.1120]$} \\
Qwen3-0.6B & GRPO-H & 1 & 0.0000 & 0.7531 & $+0.7531$ & -- \\
\midrule
Phi-3.5-mini & GRPO   & 6 & 0.0370 & 0.1620 & $+0.1249$ & $[-0.0741,+0.3768]$ \\
             & GRPO-H & 4 & 0.5701 & 0.1310 & $-0.4391$ & $[-0.8125,-0.1131]$ \\
\bottomrule
\end{tabular}
\end{table}

\begin{table}[htbp]
% Original groups: (a) updated ExPPO with F-GRPO/REPO-R, now in tab:multi-answer;
% (b) earlier ExPPO with GRPO-UR, retained here as a distinct evaluation.
\caption{Qwen2.5-7B controlled-task comparison with GRPO-UR (Rewarding the Unlikely) at the last checkpoint ($K=128$; 64 prompts). A: Accuracy: verified-response share (0--100); Solved: prompts with a success. Observed/effective yields: prompt means of distinct correct modes/exponentiated mode entropy, with zero for unsolved prompts. Correct/prompt: total verified successes divided by 64, including unsolved prompts. Bold: higher mean. B: Shared: prompts with at least two successes per method; $D$: mean distinct-mode probability of correct pairs on that set; $\Delta D$: ExPPO minus baseline; in probability units. ExPPO uses the earlier evaluation; the updated F-GRPO/REPO-R comparisons appear in \cref{tab:multi-answer}.}
\label{tab:multi-answer-nearest}
\centering
\small
\setlength{\tabcolsep}{3.5pt}
\begin{tabular*}{\linewidth}{@{\extracolsep{\fill}}lrrrrr@{}}
\toprule
\multicolumn{6}{@{}l}{\textbf{A. Fixed-budget correctness and correct-mode yield}} \\
Method & Accuracy & Solved / 64 & \shortstack[r]{Observed yield} & \shortstack[r]{Effective yield} & \shortstack[r]{Correct / prompt} \\
%\multicolumn{6}{@{}l}{\textbf{A. Fixed-budget correctness and correct-mode yield}} \\
%Model & Method & Accuracy & Solved / 64 & \shortstack[r]{Observed yield} & \shortstack[r]{Effective yield} & \shortstack[r]{Correct / prompt} \\
\midrule
% Appendix audit 2026-09-08: group (a) duplicates tab:multi-answer, Panel A.
% \multirow{3}{*}{(a)}
% & F-GRPO & \textbf{35.51} & 64 & 3.2344 & 2.3891 \\
% & REPO-R & 15.42 & 64 & 3.5781 & 2.8203 \\
% & \textbf{ExPPO} & 33.50 & 64 & \textbf{4.1719} & \textbf{3.0126} \\
% \midrule
GRPO-UR & 17.77 & 64 & 3.0000 & 2.3021 & 22.75 \\
\textbf{ExPPO} & \textbf{31.84} & 64 & \textbf{3.9688} & \textbf{2.7277} & \textbf{40.75} \\
\bottomrule
\end{tabular*}

\vspace{0.55em}
\begin{tabular*}{\linewidth}{@{\extracolsep{\fill}}lrrrrr@{}}
\toprule
\multicolumn{6}{@{}l}{\textbf{B. Conditional correct-pair diversity}} \\
Baseline & Shared / 64 & Baseline $D$ & ExPPO $D$ & $\Delta D$ \\
\midrule
% Appendix audit 2026-09-08: these pairs duplicate tab:multi-answer, Panel B.
% (a) & F-GRPO & 63 & 0.4653 & 0.5628 & $+0.0974$ & $[+0.0303,+0.1651]$ \\
% (a) & REPO-R & 63 & 0.5612 & 0.5628 & $+0.0016$ & $[-0.0758,+0.0778]$ \\
GRPO-UR & 63 & 0.4545 & 0.5113 & $+0.0569$ &\\
\bottomrule
\end{tabular*}
\end{table}

% Appendix audit 2026-09-08: the F-GRPO/REPO-R interpretation repeats sec:controlled.
% \paragraph{Broader correct support extends beyond higher accuracy.}
% ExPPO has higher observed and effective mode yields than each added baseline. F-GRPO attains higher accuracy, while ExPPO's positive paired diversity interval shows broader support among correct responses. Against REPO-R and GRPO-UR, ExPPO's yield gains accompany diversity intervals spanning zero, separating fixed-budget mode discovery from conditional diversity.
\paragraph{Higher mode yield also holds against GRPO-UR.}
ExPPO raises accuracy and both correct-mode yields over GRPO-UR. 
%The paired diversity interval spans zero, distinguishing its fixed-budget discovery gain from the conditional-diversity comparison.

\FloatBarrier
\subsection{Common-Checkpoint Continuation and Normalized Credit Allocation}
\label{app:continuation-audit}

\paragraph{Continuation from a shared trained policy.}
We evaluate one-update continuations of Qwen2.5-7B on the controlled coin-vector task. GRPO, ExPPO, and GRPO-H start from the identical ExPPO step-900 checkpoint. The study uses 32 update prompts and 32 held-out prompts. Held-out evaluation uses $K=128$ responses per prompt and a direct 2,048-token generation cap. Every evaluated policy reaches at least eight correct responses on each of the 32 held-out prompts. \Cref{tab:continuation-outcomes} reports held-out accuracy and conditional correct-pair diversity $D$, the probability that two correct responses represent different solutions. Paired 95\% confidence intervals use 10,000 prompt-bootstrap resamples and are unadjusted for multiple comparisons.

\begin{table}[H]
\caption{Qwen2.5-7B common-checkpoint continuation on 32 held-out prompts. All three update arms start from the same ExPPO-900 policy. Panel A reports accuracy in percent and diversity as a probability. Panel B reports ExPPO-minus-baseline differences in percentage points for both metrics.}
\label{tab:continuation-outcomes}
\centering
\small
\setlength{\tabcolsep}{5pt}
\begin{tabular*}{\linewidth}{@{\extracolsep{\fill}}lrr@{}}
\toprule
\multicolumn{3}{@{}l}{\textbf{A. Held-out policy outcomes}} \\
Policy & Accuracy (\%) & Diversity $D$ \\
\midrule
Shared ExPPO-900 checkpoint & 33.74 & 0.6404 \\
\quad + GRPO update & 35.64 & 0.6135 \\
\quad + ExPPO update & 36.52 & 0.6208 \\
\quad + GRPO-H update & 34.45 & 0.6001 \\
\bottomrule
\end{tabular*}

\vspace{0.6em}
%\begin{tabular*}{\linewidth}{@{\extracolsep{\fill}}lrr@{}}
%\toprule
%\multicolumn{3}{@{}l}{\textbf{B. Paired differences (percentage points)}} \\
%Comparison & Accuracy [95\% CI] & Diversity [95\% CI] \\
%\midrule
%ExPPO $-$ GRPO & $+0.88\;[-0.42,+2.32]$ & $+0.73\;[-2.04,+3.32]$ \\
%ExPPO $-$ GRPO-H & $+2.08\;[+0.59,+3.59]$ & $+2.07\;[-0.08,+4.18]$ \\
%\bottomrule
%\end{tabular*}

\begin{tabular*}{\linewidth}{@{\extracolsep{\fill}}lrr@{}}
\toprule
\multicolumn{3}{@{}l}{\textbf{B. Paired differences (percentage points)}} \\
Comparison & Accuracy & Diversity \\
\midrule
ExPPO $-$ GRPO & $+0.88$ & $+0.73\;$ \\
ExPPO $-$ GRPO-H & $+2.08\;$ & $+2.07\;$ \\
\bottomrule
\end{tabular*}

\end{table}

All three updates increase observed held-out accuracy and reduce conditional diversity relative to the shared checkpoint. ExPPO has the highest post-update point estimates for both metrics and the smallest observed diversity decrease. The paired diversity intervals span zero. These outcomes describe local continuation behavior; \cref{sec:controlled} evaluates the final policies produced by the complete training schedules.

\paragraph{Credit allocation after normalization.}
We audit the final sequence advantages $\widetilde A_i$ from \cref{eq:normalization}. For a response category $\mathcal C$, the recorded change is
\[
\Delta_{\mathcal C}
=\sum_{i\in\mathcal C}(\widetilde A_i-A_i).
\]
The sum runs over the audited update responses and measures scalar credit before token aggregation. High- and low-surprisal categories use $u_i>0$ and $u_i<0$, respectively, with $u_i$ defined by \cref{eq:rarity}. \Cref{tab:normalized-credit-audit} shows that all four aggregate directions persist after the shared normalization.

\begin{table}[H]
\caption{Post-normalization coefficient audit on the Qwen2.5-7B update responses. Values are category sums of $\widetilde A_i-A_i$ in sequence-advantage units. Surprisal classes are relative to each prompt group. Allocation effects describe the aggregate scalar coefficients used by the policy loss.}
\label{tab:normalized-credit-audit}
\centering
\small
\setlength{\tabcolsep}{4pt}
\begin{tabular*}{\linewidth}{@{\extracolsep{\fill}}llrl@{}}
\toprule
Response & Surprisal class & $\Delta_{\mathcal C}$ & Allocation effect \\
\midrule
Correct & High ($u_i>0$) & $+6.245$ & Stronger reinforcement \\
Correct & Low ($u_i<0$) & $-3.113$ & Reduced reinforcement \\
Incorrect & High ($u_i>0$) & $+8.690$ & Softer penalty \\
Incorrect & Low ($u_i<0$) & $-5.557$ & Stronger penalty \\
\bottomrule
\end{tabular*}
\end{table}

The displayed category totals give signed changes of $+3.132$ for correct responses and $+3.133$ for incorrect responses. With verifier polarity preserved, the corresponding absolute-advantage-mass change over these four categories is approximately $3.132-3.133=-0.001$, consistent with shared normalization at the reported precision. Their signed sum changes by $+6.265$. This audit characterizes the executed coefficient allocation; 
%\cref{app:limitations} discusses the semantic-rarity and utility-alignment diagnostics.

\clearpage
\subsection{Matched-Input OOD Representations}
\label{app:ood-representations}

\begin{figure}[H]
    \centering
    \includegraphics[width=\textwidth]{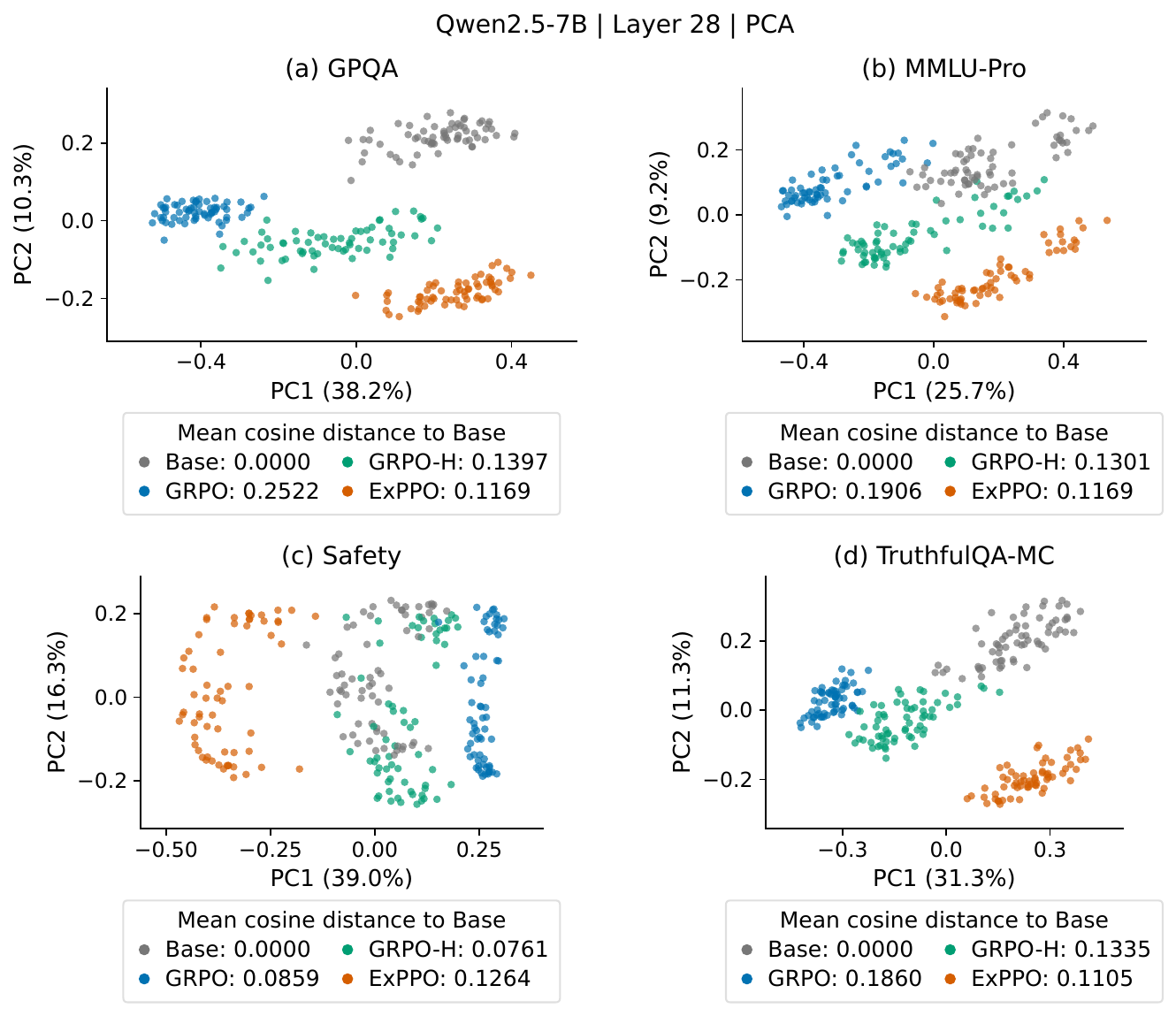}
    \caption{Qwen2.5-7B matched-input OOD representations. Each panel shows 64 shared questions for Base and three step-900 policies in a separately fitted PCA coordinate system. Axes report explained variance; legends report full-benchmark, original-space cosine distances to Base, not distances in the projection.}
    \label{fig:qwen25-ood-pca}
\end{figure}

\begin{figure}[!htbp]
    \centering
    \includegraphics[width=\textwidth]{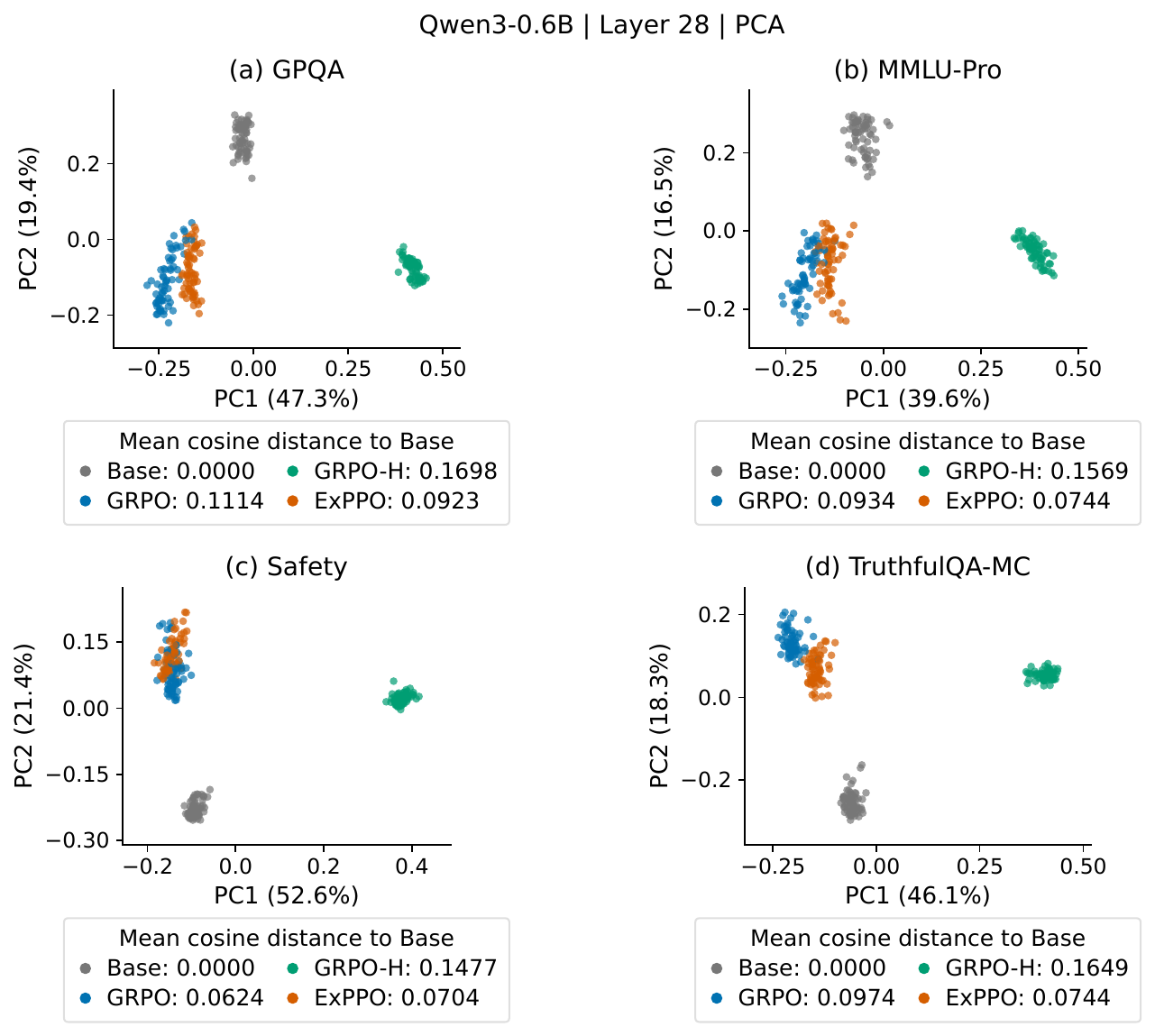}
    \caption{Qwen3-0.6B matched-input OOD representations, using the protocol in \cref{app:ood-representations}. Each panel has its own PCA coordinates and 64 shared questions across four checkpoints. Legend values measure full-benchmark cosine distance to Base before projection; the relative ordering is sensitive to answer format.}
    \label{fig:qwen3-ood-pca}
\end{figure}

\begin{figure}[!htbp]
    \centering
    \includegraphics[width=\textwidth]{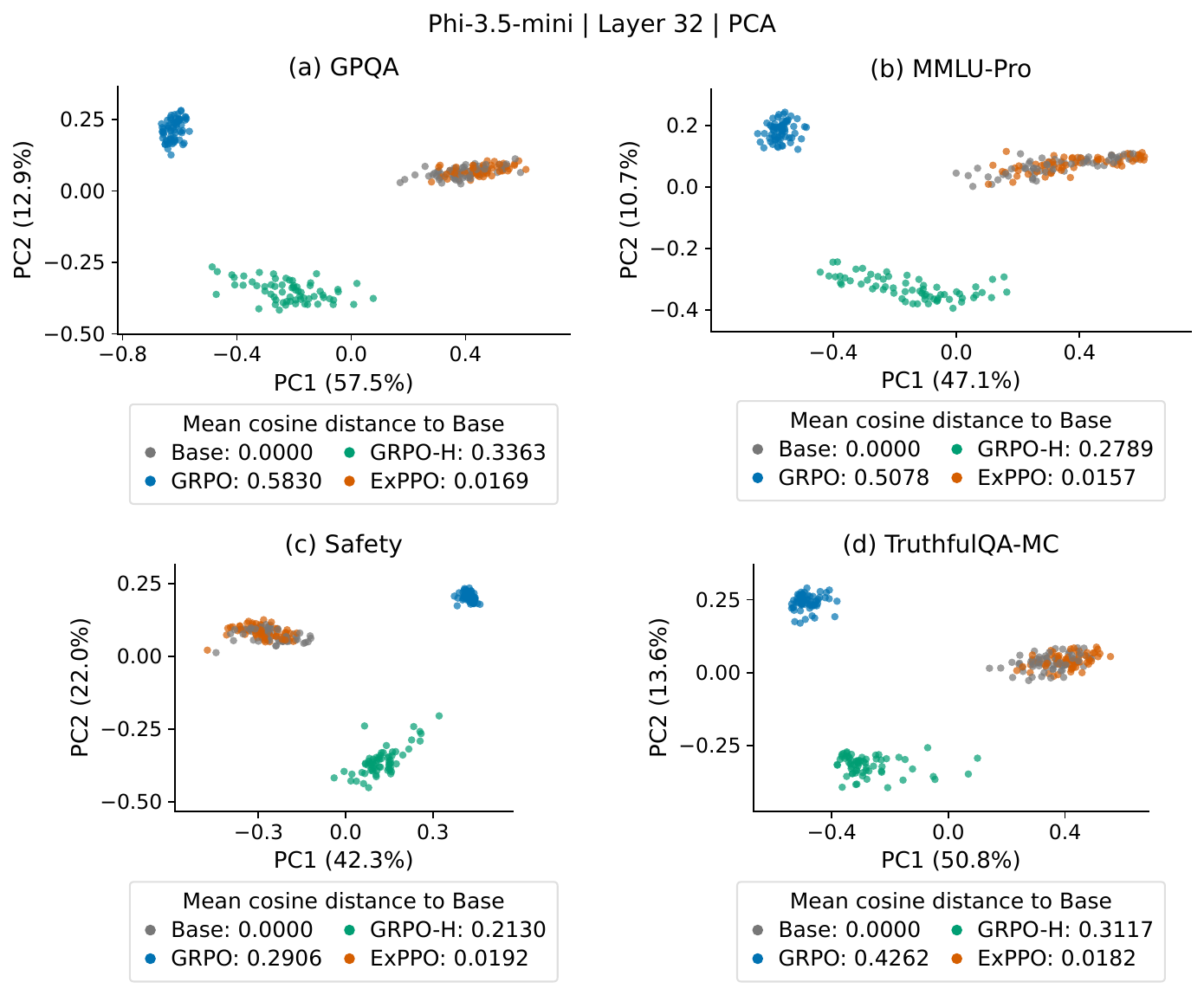}
    \caption{Phi-3.5-mini matched-input OOD representations, using the protocol in \cref{app:ood-representations}. Each panel has its own PCA coordinates and 64 shared questions across four checkpoints. Legends report full-benchmark cosine distances to the released Base reference, whose identity with the training initialization is unverified. Proximity does not establish preserved accuracy or diversity.}
    \label{fig:phi35-ood-pca}
\end{figure}

We compare Base with step-900 GRPO, GRPO-H, and ExPPO on GPQA (448 questions), MMLU-Pro (70), Safety (1,000), and TruthfulQA-MC (684). Each checkpoint receives identical questions and teacher-forced reference answers containing only the boxed correct option label. We average final-layer hidden states over answer tokens (layer 28 for Qwen; layer 32 for Phi). Within each backbone--benchmark panel, PCA is jointly fitted to the $\ell_2$-normalized representations of the same 64 questions from all four checkpoints. \Cref{fig:qwen25-ood-pca,fig:qwen3-ood-pca,fig:phi35-ood-pca} show the first two components, with explained variance on the axes. Legend values report mean cosine distances to Base over the complete benchmark, computed before dimensionality reduction. These measure representation proximity, not accuracy, diversity, or policy KL; the projection can omit differences.

ExPPO has the smallest mean cosine distance among trained methods in 10 of 12 backbone--benchmark comparisons, excepting both Qwen Safety panels. This result is specific to the boxed-label probe: full-answer-text controls change the ExPPO--GRPO ordering for Qwen3. For Phi, Base is the released reference model; its identity with the original training initialization has not been established.

\clearpage

\clearpage

\end{document}

%% file: math_commands.tex
\usepackage{amsmath,amsfonts,bm}

\def\eqref#1{equation~\ref{#1}}
\def\1{\bm{1}}

\DeclareMathAlphabet{\mathsfit}{\encodingdefault}{\sfdefault}{m}{sl}
\SetMathAlphabet{\mathsfit}{bold}{\encodingdefault}{\sfdefault}{bx}{n}

\newcommand{\E}{\mathbb{E}}

\newcommand{\Var}{\mathrm{Var}}

\newcommand{\Cov}{\mathrm{Cov}}